\documentclass{article}

\usepackage{balance} 

\usepackage[utf8]{inputenc} 
\usepackage[T1]{fontenc}    
\usepackage{hyperref}       
\usepackage{url}            
\usepackage{booktabs}       
\usepackage{amsfonts}       
\usepackage{nicefrac}       
\usepackage{microtype}      
\usepackage{tabularx}

\usepackage{mathtools}
\usepackage{caption}
\usepackage{subcaption}
\usepackage{amsfonts}
\usepackage{mathrsfs}

\usepackage{graphicx}

\usepackage[accepted]{icml2021}

\icmltitlerunning{Balancing Rational and Other-Regarding Preferences in Cooperative-Competitive Environments}

\begin{document}

\twocolumn[
\icmltitle{Balancing Rational and Other-Regarding Preferences in Cooperative-Competitive Environments}

\icmlsetsymbol{equal}{*}

\begin{icmlauthorlist}
\icmlauthor{Dmitry Ivanov}{equal,hse,jb}
\icmlauthor{Vladimir Egorov}{equal,hse,jb}
\icmlauthor{Aleksei Shpilman}{hse,jb}
\end{icmlauthorlist}

\icmlaffiliation{hse}{HSE University, Russian Federation}
\icmlaffiliation{jb}{JetBrains Research}

\icmlcorrespondingauthor{Dmitry Ivanov}{diivanov@hse.ru}
\icmlcorrespondingauthor{Vladimir Egorov}{vsegorov@edu.hse.ru}


\vskip 0.3in
]

\printAffiliationsAndNotice{\icmlEqualContribution}

\begin{abstract}
Recent reinforcement learning studies extensively explore the interplay between cooperative and competitive behaviour in mixed environments. Unlike cooperative environments where agents strive towards a common goal, mixed environments are notorious for the conflicts of selfish and social interests. As a consequence, purely rational agents often struggle to achieve and maintain cooperation. A prevalent approach to induce cooperative behaviour is to assign additional rewards based on other agents' well-being. However, this approach suffers from the issue of multi-agent credit assignment, which can hinder performance. This issue is efficiently alleviated in cooperative setting with such state-of-the-art algorithms as QMIX and COMA. Still, when applied to mixed environments, these algorithms may result in unfair allocation of rewards. We propose BAROCCO, an extension of these algorithms capable to balance individual and social incentives. The mechanism behind BAROCCO is to train two distinct but interwoven components that jointly affect each agent’s decisions. Our meta-algorithm is compatible with both Q-learning and Actor-Critic frameworks. We experimentally confirm the advantages  over the existing methods and explore the behavioural aspects of BAROCCO in two mixed multi-agent setups.  
\end{abstract}

\section{Introduction}\label{sec-introduction}

Human cooperation is considered an evolutionary puzzle in the economic literature \cite{Axelrod1390, FehrInequity,johnson2003puzzle,colman2006puzzle,rand2013human}. Despite the predictions of the rational choice theory to act selfishly \cite{scott2000rational}, people of different age, gender, culture, and socioeconomic status engage into cooperation in a multitude of economic situations \cite{croson1999gender,henrich2001cooperation,alvard2004ultimatum,benenson2007children,chen2013family,kettner2016old}. A notable example of such situations is prisoner's dilemma \cite{rapoport1965prisoner}, where a rational agent chooses to defect despite his preference of mutual cooperation over mutual defection. One of the possible mechanisms to resolve the paradox implies that the agents take social and other-regarding preferences into account during decision making \cite{FehrInequity,fehr2002social}.

The questions of emergence and maintenance of cooperation are mirrored in the Multi-Agent Reinforcement Learning (MARL) literature \cite{tan1993multi,lowe2017multi,sunehag2017value,qmix,coma,peysakhovich2018prosocial}. Numerous works have repeatedly demonstrated that purely rational agents are unable to maintain mutually beneficial cooperation, unlike the agents guided by social incentives \cite{peysakhovich2018prosocial,hughes2018inequity,jaques2019social,wang2019evolving}. Despite this, training fully social agents can be undesirable when fairness is a concern.

\begin{figure*}[h]
\centering
\begin{subfigure}{.462\linewidth}
\centering
\includegraphics[width=\linewidth]{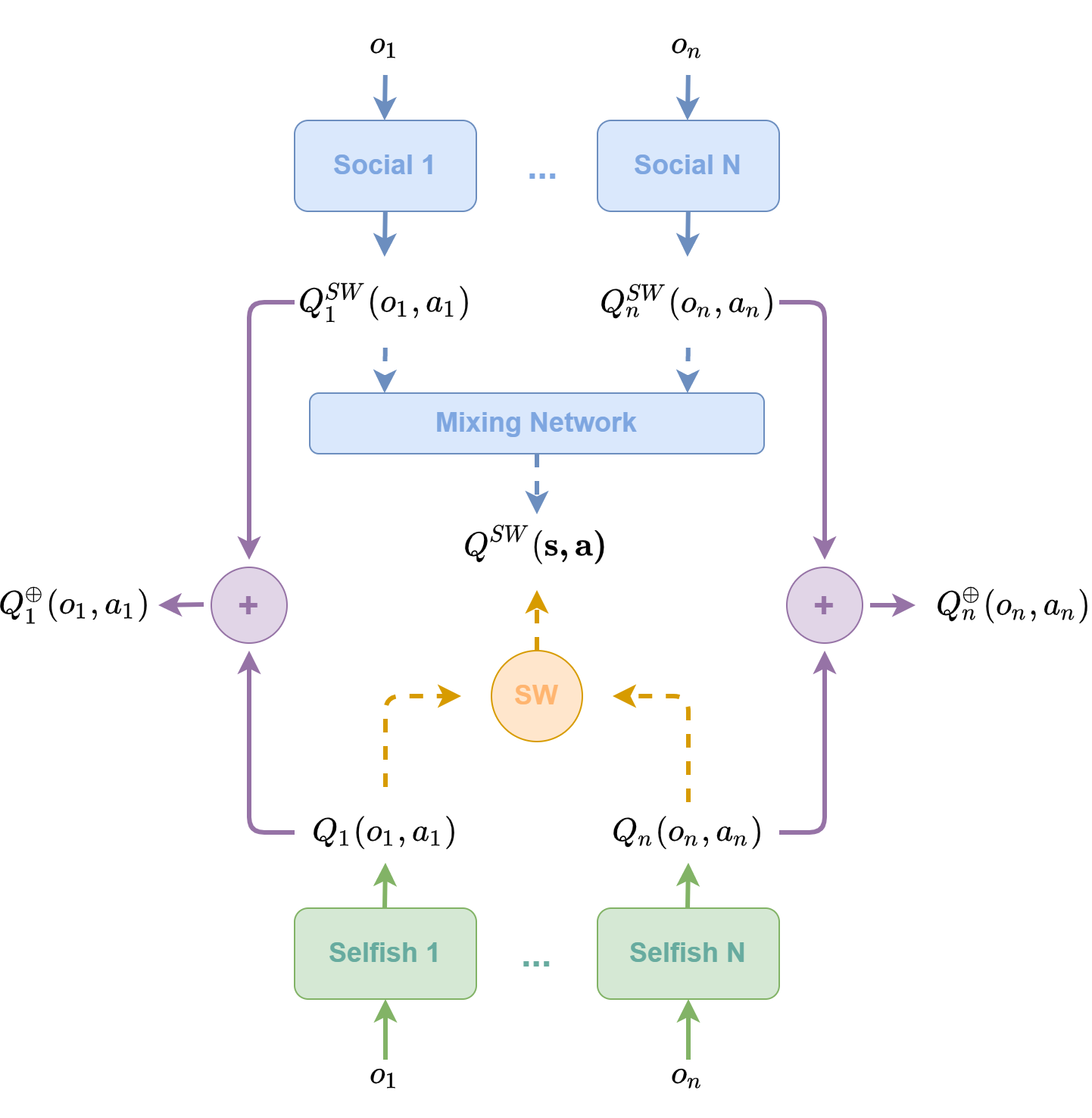}
\caption{BAROCCO in Q-learning framework}
\end{subfigure}%
\begin{subfigure}{.53\linewidth}
\centering
\includegraphics[width=\linewidth]{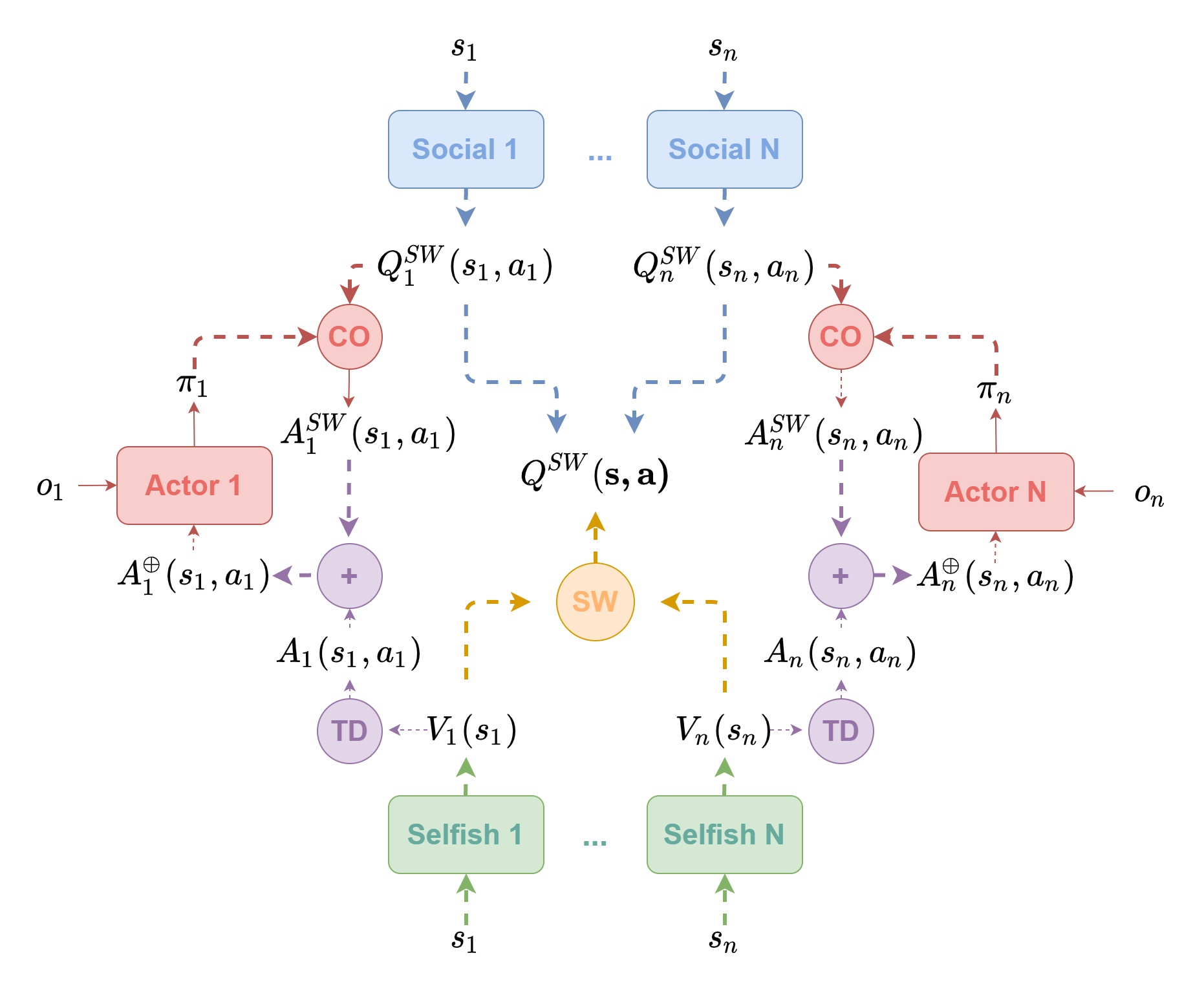}
\caption{BAROCCO in Actor-Critic framework}
\end{subfigure}
\caption{BAROCCO. Solid lines represent parts that are used during both training and execution. Dashed lines represent parts that are only used during training. a) Selfish components predict selfish Q-values $Q_i$ and are trained independently. Social components predict per-agent contributions $Q_i^{SW}$, combined into social Q-value $Q^{SW}$ through mixing network. Social components and mixing network are trained end-to-end to approximate temporal difference target of social welfare (SW), defined as a combination (e.g. sum) of $Q_i$. Agents act according to combined Q-values $Q_i^\oplus$, which are convex mixtures of $Q_i$ and $Q_i^{SW}$. b) Selfish components predict selfish values $V_i$ and are trained independently. Social components predict social Q-values $Q_i^{SW}$ and are trained to approximate temporal difference target of social welfare (SW), defined as a combination (e.g. sum) of $V_i$. Selfish advantages $A_i$ are estimated as temporal differences (TD) of $V_i$. Social advantages $A_i^{SW}$ are estimated by subtracting counterfactual (CO) baselines from $Q_i^{SW}$. Decentralized policies $\pi_i$ are trained via policy gradient on combined advantages $A_i^\oplus$, which are convex mixtures of $A_i$ and $A_i^{SW}$.}
\label{fig:barocco}
\end{figure*}


As an example, consider the problem of coordination of autonomous vehicles. On the one hand, each car's passengers have their own goals in terms of destination and desirable arrival time. Treating this problem as fully cooperative, as implied in \cite{cao2012overview,qmix}, may favor solutions where agents sacrifice these personal goals for the social good. For instance, a fully cooperative agent would be willing to let the other cars pass and stay on a crossroad indefinitely as long as average arrival time decreases. In contrast, a selfish agent would not. This example illustrates how fairness emerges from selfishness. On the other hand, it is still crucial for each agent to avoid creating inconvenient or dangerous situations for other cars. Therefore, this scenario falls in-between selfish and social and requires agents to balance these preferences.

The simplest way to achieve such balance is to train agents on a mixture of selfish and social rewards \cite{durugkar2020balancing}, which we refer to as Cooperative Reward Shaping (CRS). In this work, we define selfish reward as the standard reward an agent receives in the environment, and social reward as some combination (e.g. sum) of selfish rewards of all agents. However, CRS implies decentralized training and does not address several crucial issues of MARL, such as credit assignment, partial observability, and inherent non-stationarity \cite{agogino2004unifying,non-stationarity,hernandez2019survey}. The two latter issues can be alleviated by considering global information and actions of other agents during training, as done in MADDPG \cite{lowe2017multi}. Still, the combination of CRS and MADDPG does not address credit assignment of agents to the social welfare. On the other hand, all these issues are addressed by the techniques from fully cooperative MARL like QMIX \cite{qmix} or COMA \cite{coma} that were shown to outperform decentralized training in such complex environments as StarCraft 2 \cite{vinyals2017starcraft}. Still, these techniques are only concerned with team performance and ignore fairness. 

In this paper we propose a meta-algorithm that extends techniques like QMIX and COMA to mixed environments with capability to balance the incentives. We refer to this meta-algorithm as BAROCCO, i.e. BAlancing Rational and Other-regarding preferences in Cooperative-COmpetitive environments. BAROCCO is based on the insight that instead of relying on a single model to balance incentives via CRS, two distinct components, i.e. selfish and social, can be trained concurrently and combined during decision making. While we show that mathematically the two approaches are equivalent, the latter approach allows us to train the social component via techniques that address credit assignment.

More specifically, BAROCCO is compatible with both Q-learning and Actor-Critic frameworks. In the case of Q-learning framework, for each agent we train selfish Q-value via Rainbow \cite{rainbow} and social Q-value via QMIX \cite{qmix}. During decision making, the agents choose the action that maximizes the mixture of the two Q-values, and the importance of each Q-value is controlled via predefined prosociality coefficient. In the case of Actor-Critic framework, we train selfish critic via a variant of MADDPG \cite{lowe2017multi} and social critic via COMA \cite{coma}. Then, the actor is trained via proximal policy gradient \cite{schulman2017proximal} using a mixture of predictions of these two critics.

For both frameworks, we show that varying the prosociality coefficient in BAROCCO results in trade-off of efficiency and fairness. In particular, we find that fully social agents may choose to concentrate all environment's rewards in one particular agent, whereas agents with a non-zero selfish component refuse to make such sacrifices. More surprisingly, in some cases we find that less social agents are not only more fair but also more efficient.

A crucial novelty of BAROCCO concerns the training of the social component. The natural approach would be to construct common reward as a combination of selfish rewards. Instead, we directly combine selfish values, omitting construction of common reward. We respectively refer to these approaches as short-term and long-term. While in certain cases the two approaches are mathematically equivalent, the long-term approach might be more suitable for mixed environments. We also formulate two qualitative advantages of the long-term approach: compatibility with a broader set of social welfare functions and applicability to a wider range of environments.

Finally, an alternative to achieving fairness through selfishness could be to train a fair centralized system by maximizing minimum of agents' payoffs rather than sum. We show that such procedure can also be viable, but only if the system is trained via the long-term approach used in BAROCCO. In this case, the selfish components are vital for efficiency, albeit are only used to estimate target and do not influence agents' decisions directly.

\section{Definitions and Background}\label{sec-background}


\subsection{Notations}\label{sec-background-markov}

A tuple $\mathcal{G} = (S, N, \mathcal{A}, O, T, r)$ defines a temporally-extended Markov game \cite{littman1994markov}, where:

\begin{itemize}
    \item Let $S$ be set of states $s$, $N$ be number of players, $\mathcal{A}_i$ be set of actions $a$ of player $i$. Let $s_i = (s, a_{-i})$ denote concatenation of state $s$ and actions of other agents $a_{-i}$
    \item Let $\mathcal{O}: S \times N \rightarrow \mathbb{R}^d$ be function that specifies d-dimensional observations available to each agent. Let $O_i$ be set of observations $o_i$ of agent $i$.
    \item Let $T: S \times \mathcal{A}_1 \times \dots \times \mathcal{A}_N \rightarrow \Delta(S)$ be transition function, where $\Delta$ is set of discrete probability distributions over $S$. Let $\Delta_0(S)$ be the distribution of initial states. 
    \item Let $r_i: S \times \mathcal{A}_1 \times \dots \times \mathcal{A}_N \rightarrow \mathbb{R}$ be reward function for each player $i$. 
    \item Let $R_i(s) = \overset{\infty}{\underset{t=0}{\sum}} \left[ \gamma^t r_i(s_t, a_{1_t}, \dots, a_{N_t}) \mid s_0 = s \right]$ be return of player $i$ in state $s$ with discount factor $\gamma \in [0, 1)$. Let $\pi_i: O_i \rightarrow \Delta(\mathcal{A}_i)$ be policy of player $i$. Let $\pi_i(a \mid o_i)$ be probability of taking action $a$ in local state $o_i$.
    \item Let $V_{i}(s) = \mathop{\mathbb{E}}_{\pi_i}[R_i(s)]$ be state value function, $Q_{i}(s,a) = \mathop{\mathbb{E}}_{\pi_i}[R_i(s) \mid a_{i_0} = a]$ be state-action value function, $A_{i}(s, a) = Q_{i}(s, a) - V_{i}(s)$ be advantage function. Subscript $t$ will denote time-step, e.g. $V_{i_t} = V_{i}(s_t)$. Bold font will denote vector, e.g. $\textbf{a} = (a_1, \dots, a_N)$, $\textbf{V}(s) = (V_1(s), \dots, V_N(s))$.
    \item Let $SW: \mathbb{R}^N \rightarrow \mathbb{R}$ be social welfare function that evaluates well-being of all agents. The simplest example of its application is sum of agents' rewards: $SW(\textbf{r}) = \sum_i{r_i}$.
\end{itemize}

\subsection{Single-Agent Reinforcement Learning}\label{sec-background-rl}


\paragraph{Deep Q-Learning.}

In Q-learning \cite{watkins1992q}, the agent's goal is to learn Q-values for each state-action pair, and the agent's policy is to choose actions that correspond to the highest Q-values. This approach has been successfully applied to such complex environments as Atari games when coupled with deep learning \cite{rainbow,badia2020agent57}. In Deep Q-Networks (DQN) \cite{DQN}, the Q-values are no longer tabular and are instead approximated with a neural network trained with squared Temporal Difference (TD) loss function: 

\begin{equation}\label{eq:loss_dqn}
\mathcal{L}_{TD}(Q_t, y_t) = (y_t - Q_t)^2 
\end{equation}

where $y_t = r_t + \gamma \max_{a_{t+1}'}Q_{t+1}$. The essential features of DQN are a replay buffer, which enables the reuse of past experiences, and a separate network for target estimation, which stabilises training. The performance of DQN was greatly improved in Rainbow \cite{rainbow} by combining several modifications proposed in different papers.

\paragraph{Actor-Critic.}

In Actor-Critic framework \cite{A3C}, the Actor's goal is to learn a policy $\pi(s)$ that maximizes agent's long-term payoffs predicted by the Critic. A widely-used method is proximal policy optimization (PPO) \cite{schulman2017proximal}, where the Actor's neural network is trained on the following loss: 

\begin{equation}\label{eq:loss_ppo}
\mathcal{L}_\pi(A_t) = -\min(\mathcal{R}_t A_t,  clip(\mathcal{R}_t, 1 - \epsilon, 1 + \epsilon) A_t)
\end{equation}

where  $\mathcal{R}_t = \frac{\pi(a_t | s_t)}{\pi_{old}(a_t | s_t)}$ denotes probability ratio of the policies after and before the update. Using this loss ensures that the agent's policy stays within a trust region during the update. The advantage is defined as $A_t = y_t - V_t$, where $y_t = r_t + \gamma V_{t+1}$. The Critic's neural network is independently trained to predict $V_t$ by minimizing squared TD error $\mathcal{L}_{TD}(V_t, y_t)$. 


\subsection{Independent and Centralized Multi-Agent Reinforcement Learning}\label{sec-background-independent}

In Multi-Agent Reinforcement Learning (MARL), multiple agents learn and interact in the same environment. One of the simplest approaches in MARL is to train agents independently using unmodified single-agent RL techniques \cite{tan1993multi}. Unfortunately, this naive approach invalidates convergence guarantees \cite{lowe2017multi} of Q-Learning \cite{watkins1992q} and Actor-Critic \cite{konda2000actor}. The reason for that is the inherent non-stationarity of multi-agent environments \cite{laurent2011world,non-stationarity}. Furthermore, independent MARL does not address the issue of credit assignment in environments with common reward \cite{wolpert2002optimal,agogino2004unifying}. Nevertheless, this approach can be effective in both cooperative \cite{berner2019dota} and mixed \cite{leibo2017SSD,tampuu2017multiagent} setups. As the opposite extreme, the fully centralized approach reduces MARL to single-agent RL by controlling all agents simultaneously based on global information. Unfortunately, centralized MARL suffers from scalability issues due to exponential growth of the joint action space in the number of agents \cite{guestrin2002multiagent,sunehag2017value}. 

\subsection{Centralized Training with Decentralized Execution}\label{sec-background-ctde}


Centralized Training with Decentralized Execution (CTDE) is a compromise between independent and centralized MARL \cite{kraemer2016multi,lowe2017multi,sunehag2017value,qmix,coma,son2019qtran}. Under this paradigm, training can be enhanced with the use of global information as long as it results in decentralized policies. Typically, CTDE techniques alleviate the issues of multi-agent credit assignment and/or non-stationarity while effectively dealing with the curse of dimensionality.

QMIX \cite{qmix} is an algorithm designed to train multiple Q-learning agents in cooperative environments. During training, it approximates the joint Q-value as a monotonic mixture of the individual Q-values: $Q(s, \textbf{a}) = \mathcal{M}(Q_1(o_1, a_1), \dots, Q_N(o_N, a_N))$. During execution, each agent acts according to its individual Q-value $Q_i(o_i, a_i)$, restricting the use of global information to the training phase. The function $\mathcal{M}$ is trained as a mixture network in an end-to-end fashion via TD loss $\mathcal{L}_{TD}(Q(s_t, \textbf{a}_t), r_t + \gamma \max_{\textbf{a}_{t+1}'} Q(s_{t+1}, \textbf{a}_{t+1}'))$. By enforcing the monotonicity of $\mathcal{M}$, the joint Q-value can be factorised in a way that preserves the order of actions. As a result, maximization over the joint action becomes tractable: $max_{\textbf{a}}Q(s, \textbf{a}) = \mathcal{M}(max_{a_1}Q_1(o_1, a_1), \dots, max_{a_N}Q_N(o_N, a_N))$. To utilize global information $s$, the weights of the mixture network are predicted with a set of hypernetworks \cite{hypernetworks}. QMIX is a direct extension of Value Decomposition Networks \cite{sunehag2017value}, where the joint Q-value is simply approximated as a sum of agents contributions $Q_i$ rather than a monotonic mixture.  

Counterfactual Multi-Agent policy gradient (COMA) \cite{coma} is an adaptation of Actor-Critic framework to cooperative environments. COMA uses an efficiently designed centralized critic, which outputs Q-values $Q_i(s, a_i \mid a_{-i})$ for a specified agent $i$ based on the global state $s$ and the actions of the other agents $a_{-i}$. Furthermore, COMA estimates advantage for each agent by marginalising out the agent's action while keeping actions of other agents fixed: $A_i(s, a_i \mid a_{-i}) = Q_i(s, a_i \mid a_{-i}) -  \sum_{a_i'} \pi_i(a_i' \mid o_i) Q_i(s, a_i' \mid a_{-i})$. Decentralized policies $\pi_i(o_i)$ are trained on these advantages via policy gradient.

Multi-Agent Deep Deterministic Policy Gradient (MADDPG) \cite{lowe2017multi} is a CTDE algorithm specifically designed for mixed environments. The core idea is to train DDPG agents using centralized critics conditioned on global state $s$ and actions of all agents $\textbf{a}$. Similarly to the enhanced critic in COMA, this modification reduces variance of policy gradient, as well as addresses non-stationarity and partial observability. However, MADDPG does not concern credit assignment. In our paper, we apply the same centralization of critic as in MADDPG to multi-agent PPO with discrete action spaces when training the selfish components of Actor-Critic agents.

\subsection{Cooperative Reward Shaping}\label{sec-background-crs}

We broadly define Cooperative Reward Shaping (CRS) as reward shaping with respect to the behaviour of other agents, e.g. their rewards \cite{lerer2017tft,peysakhovich2018consequentialist,peysakhovich2018prosocial,hughes2018inequity,wang2019achieving}, temporal differences \cite{hostallero2018ped}, policies \cite{jaques2019social}, etc. CRS aims to learn cooperative yet not selfless policies in mixed environments. In this paper, we will only be concerned with a particular instance of CRS where agents' rewards are mixed:

\begin{equation}\label{eq:crs}
    r_i^\oplus = (1 - \lambda) r_i + \lambda SW(\textbf{r})
\end{equation}



\noindent where $\lambda$ is prosociality coefficient and $r_i^\oplus$ is combined reward. The agents are fully selfish when $\lambda = 0$ and fully social when $\lambda = 1$. The social reward defined as a sum of individual rewards is routinely used in MARL papers to train cooperative policies \cite{lerer2017tft,peysakhovich2018consequentialist,peysakhovich2018prosocial,wang2019achieving}. The idea to train agents on a convex mixture of selfish and social rewards similar to (\ref{eq:crs}) is explored by \citet{durugkar2020balancing}. While being simple, this approach is limited in its incapability to find some of the Pareto optimal solutions, particularly the solutions that lie in concave regions of the Pareto front \cite{vamplew2008limitations}.

\section{BAROCCO}\label{sec-barocco} 

\subsection{Factorization of CRS}\label{sec-barocco-crs}

While CRS can achieve balance between selfish and social incentives, it does not address multi-agent credit assignment, which can be crucial for performance. At the same time, CTDE algorithms like QMIX and COMA address credit assignment but are intended for cooperative environments, requiring agents to forgo selfish incentives. As a middle-ground, we notice that the value $V_i^{\oplus}$ that CRS agents optimize can be factored as a mixture of selfish and social values $V_i$ and $V^{SW}$ respectively:


\begin{multline}
    \label{eq-factorization}
    V_i^{\oplus}(s) = \mathbb{E}_{\pi_i} \sum_t \gamma^t \left((1 - \lambda) r_{i_t} + \lambda SW(\mathbf{r}_t)\right) \mid s_0 = s \\ = (1 - \lambda) \mathbb{E}_{\pi_i} \sum_t \gamma^t r_{i_t} + \lambda \mathbb{E}_{\pi_i} \sum_t \gamma^t SW(\mathbf{r}_t) \mid s_0 = s \\ = (1 - \lambda) V_i(s) + \lambda V^{SW}(s)
\end{multline}

\noindent where expectations over policies of other agents $\pi_{-i}$ and over transition function $T$ are omitted for brevity. Note that the same factorization can be applied to Q-value $Q_i^{\oplus}$ and advantage $A_i^{\oplus}$. 

The factorization (\ref{eq-factorization}) allows us to train the social component separately from the selfish component via algorithms like QMIX and COMA that address credit assignment in cooperative environments. This technique forms the basis for BAROCCO. 
In Section \ref{sec-barocco-sw}, we take a more in-depth look on the social value and propose an alternative definition that is not based on common reward. Then, in Sections \ref{sec-barocco-dqn} and \ref{sec-barocco-ppo} we discuss the specifics of training and combining the two components within Q-learning and Actor-Critic frameworks. Pseudocode of BAROCCO is available in Appendix.

\subsection{Assessing Social Welfare}\label{sec-barocco-sw}

In the previous subsection, we defined social value based on common reward $SW(\textbf{r})$, which is a combination of individual rewards $r_i$ of all agents. We will denote this value as $V^{{SW}_S}$, where subscript $S$ stands for `short-term`. For convenience, this definition is repeated in (\ref{eq-sw-short}). Training COMA critic or QMIX on TD loss based on this definition of value when the agents are fully social (i.e. $\lambda = 1$) is the most straightforward way to extend these algorithms to mixed environments that will be referred to as Vanilla. 

The difference between BAROCCO and Vanilla algorithms is two-fold. First, BAROCCO agents can consider both selfish and social motives during decision making, which is also reflected in the modified training procedure. This will be discussed in details in the following subsections. Second, BAROCCO utilizes an alternative definition of social value $V^{{SW}_L}$, formulated in (\ref{eq-sw-long}) and referred to as `long-term`. Long-term value is not based on common reward $SW(\textbf{r})$ and is instead defined as a combination of agents' selfish values $V_i$. Essentially, the two values $V^{{SW}_S}$ and $V^{{SW}_L}$ differ in the order in which expectation, sum, and social welfare function are applied. We experimentally confirm that replacing $V^{{SW}_S}$ with $V^{{SW}_L}$ can increase performance. Additionally, we identify two qualitative advantages of the long-term value. We briefly formulate these advantages below and verify them experimentally in Section \ref{sec-experiments-results}. We also provide detailed examples in Appendix.


\begin{equation}
\label{eq-sw-short}
  V^{SW_S} = \mathop{\mathbb{E}}_{\mathbf{\pi}}\left[ \sum_t \left( \gamma^t SW(\textbf{r}_t) \right) \right]
\end{equation}

\begin{equation}
\label{eq-sw-long}
V^{SW_L} = SW \left(\mathop{\mathbb{E}}_{\mathbf{\pi}}\left[ \sum_t \left( \gamma^t \textbf{r}_t \right) \right] \right) = SW(\textbf{V}_t)
\end{equation}

The first limitation of $V^{SW_S}$ is in the choice of social welfare functions $SW$. When $SW$ is chosen as sum, $V^{SW_S}$ is mathematically equivalent to $V^{SW_L}$ due to commutativity of sum with expectation (although practical implementations of the algorithms still differ). However, this is not always the case. For instance, choosing $SW$ as minimum can be a way to account for both efficiency and fairness \cite{rawls2009theory}. In this case, maximizing $V^{SW_S}$ requires fair reward distribution at each time-step, whereas to maximize $V^{SW_L}$ the rewards should only be fairly distributed on average. While solving the first task is sufficient for solving the second, it is also unnecessarily constraining and might result in poor performance. Our experiments support this conjecture.

The second limitation of $V^{SW_S}$ is inapplicability to environments where trajectory lengths are variable. As an example, consider an environment where the agents receive negative rewards upon termination. In such environment, an agent that maximizes $V^{SW_S}$ might adopt two opposite strategies. The first strategy is to prolong the episodes of all agents, thus postponing the negative rewards. The second strategy is to terminate own episode early, thus avoiding the negative rewards from other agents altogether. In contrast, an agent that optimizes $V^{SW_L}$ anticipates termination of other agents regardless of witnessing it and therefore can only adopt the first strategy. This issue is akin to the bias in rewards identified in generative adversarial imitation learning \cite{kostrikov2018discriminator}. 

\begin{figure*}[h]
\centering
\begin{subfigure}{.33\linewidth}
\centering
\includegraphics[width=\linewidth]{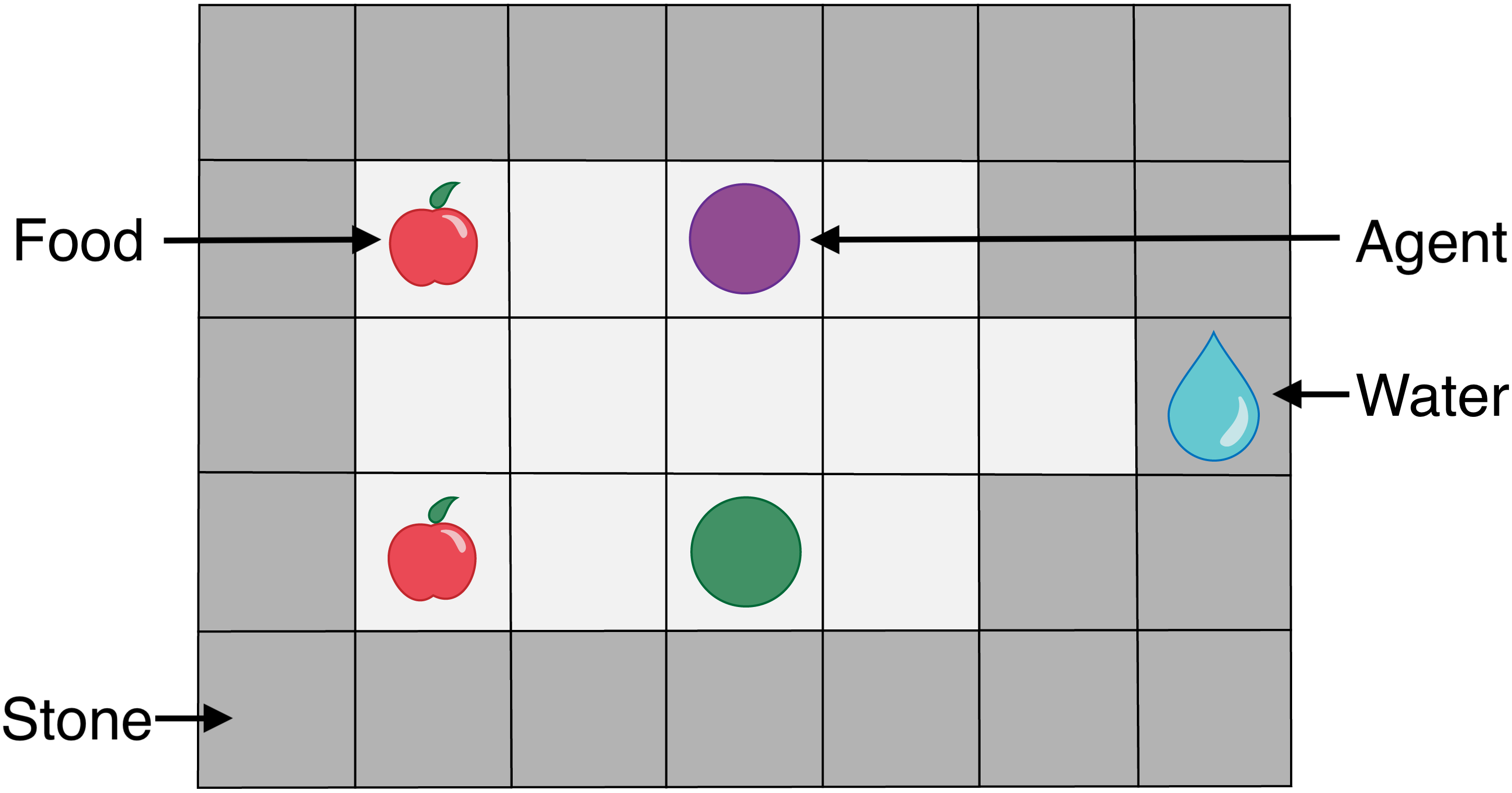}
\caption{Eldorado}
\end{subfigure}
\begin{subfigure}{.625\linewidth}
\centering
\includegraphics[width=\linewidth]{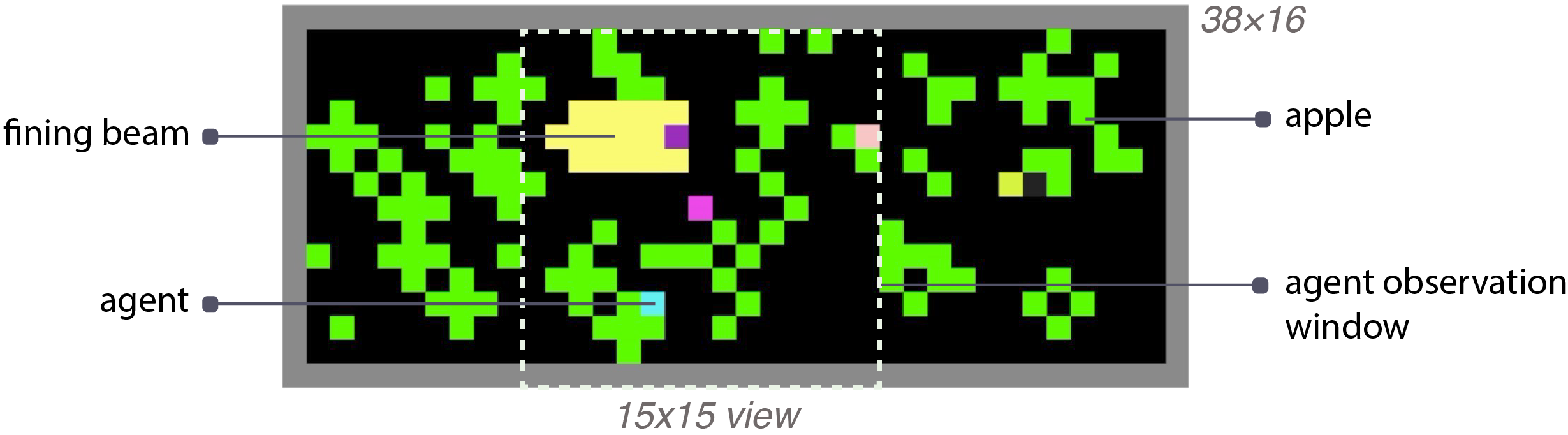}
\caption{Harvest}
\end{subfigure}
\caption{Environments. Illustration of Harvest map is taken from \cite{hughes2018inequity}.}
\label{fig:map}
\end{figure*}

As a side note, if simultaneous optimization of payoffs of multiple agents is viewed as multi-objective optimization, then the proposed long-term approach to MARL corresponds to the `scalarization of the expected return` approach to multi-objective RL \cite{roijers2013survey}. It could also be interesting to explore the alternative `expectation of the scalarized return` approach, which would imply changing the order of $SW$ function and expectation in (\ref{eq-sw-long}), but we leave this direction to the future work.

\subsection{Combining Independent DQN and QMIX}\label{sec-barocco-dqn}

Here we describe BAROCCO in Q-learning framework. The algorithm is schematically illustrated in Figure \ref{fig:barocco}a.

When choosing an action, each agent maximizes the following convex combination of Q-values:

\begin{equation}\label{eq:Q_tot}
Q_i^\oplus(o_i, a_i) = (1 - \lambda) Q_i(o_i, a_i) + \lambda Q_i^{SW}(o_i, a_i)    
\end{equation}

\noindent where $Q_i$, $Q_i^{SW}$, and $Q_i^\oplus$ denote selfish, social, and combined Q-values, respectively. Equation (\ref{eq:Q_tot}) is, in essence, equation (\ref{eq-factorization}) rewritten for Q-values, but with one distinction: the social component is not common but is based on each agent's contribution to social welfare. These contributions are disentangled via mixture network, as proposed in QMIX. Although the two Q-values $Q_i$ and $Q_i^{SW}$ are optimized separately, they still affect each other through the agent's policy.

For each agent $i$, the selfish Q-value $Q_i$ is trained via independent Q-learning (see Section \ref{sec-background-rl}). In particular, we use Rainbow architecture \cite{rainbow}, which is a modification of DQN \cite{DQN}. The only important distinction is that the agents do not act according to the estimated Q-values, i.e. they maximize $Q_i^\oplus$ rather than $Q_i$. At the same time, $Q_i$ should be the expectation over the behavioural policy $\pi_i$ according to the definition of selfish value in (\ref{eq-factorization}). To account for this discrepancy, the TD target is modified akin to double Q-learning. Specifically, maximization of $Q_i$ over actions is replaced with $Q_i$ of the action that maximizes $Q_i^{\oplus}$: 

\begin{equation}\label{eq:Q_target_selfish}
    y_{i_t}=r_{i_t} + \gamma Q_i(o_{i_{t+1}}, argmax_{a_{i_{t+1}}'} Q_i^\oplus(o_{i_{t+1}}, {a_{i_{t+1}}'}))
\end{equation}


The social component is based on QMIX. The common Q-value $Q^{SW}$ is trained on TD loss and is disentangled into agents' individual contributions $Q_i^{SW}$ via mixture network (see Section \ref{sec-background-ctde}). These individual contributions constitute social components for each agent. We explore two alternative estimates of TD target $y^{SW}$ for QMIX that correspond to two definitions of social values, discussed in Section \ref{sec-barocco-sw}. The first estimate (\ref{eq:Q_target_prosocial_vanilla}) is based on the common reward and the prediction of QMIX for the next state. As in the case of the selfish component, the target is modified with respect to the combined Q-values $Q_i^\oplus$. When $\lambda = 1$, this target is equivalent to the target used in Vanilla QMIX. The second estimate (\ref{eq:Q_target_prosocial_barocco}), used in BAROCCO, is based on TD targets for the selfish components.

\begin{equation}
\label{eq:Q_target_prosocial_vanilla}
  y_t^{SW_S}=SW(\textbf{r}_t) + \gamma Q^{SW}(s_{t+1}, argmax_{\textbf{a}'}\textbf{Q}^\oplus(\textbf{o}_{t+1}, \textbf{a}'))
\end{equation}

\begin{equation}
\label{eq:Q_target_prosocial_barocco}
y_t^{SW_L}=SW(\textbf{y}_t)
\end{equation}

In our implementation, both Vanilla QMIX and BAROCCO utilize noisy exploration \cite{noise}.

\subsection{Combining MADDPG, COMA, and PPO}\label{sec-barocco-ppo}

Here we describe BAROCCO in Actor-Critic framework. The algorithm is schematically illustrated in Figure \ref{fig:barocco}b.

Each agent acts according to its decentralized policy $\pi_i(o_i)$ trained on PPO loss $\mathcal{L}_\pi(A_i^\oplus)$, where $A_i^\oplus$ is a convex combination of selfish and social advantages $A_i$ and $A_i^{SW}$:

\begin{equation}\label{eq:A_tot}
A_i^\oplus(s_i, a_i) = (1 - \lambda) A_i(s_i, a_i) + \lambda A_i^{SW}(s_i, a_i)    
\end{equation}


For agent $i$, the selfish advantage is estimated as TD of a critic that predicts the agent's selfish value:  $A_i = y_{i_t} - V_{i_t}$, where $y_{i_t} = r_{i_t} + \gamma V_{i_{t+1}}$. The selfish critic estimates value with respect to the behavioural policy $\pi_i$, which corresponds to the definition of $V_i$ in (\ref{eq-factorization}). So, no additional modifications of its target are required. Note that instead of using only local observations, the critic makes predictions based on concatenation of global state and actions of other agents $s_i$. Therefore, it is trained with a variation of MADDPG.


The social component is based on COMA. For each agent, its social critic is trained on TD loss and predicts social Q-value $Q_i^{SW} (s_i, a_i)$. Then, the advantage $A_i^{SW} (s_i, a_i)$, i.e. the effect of the agent's actions on social welfare, is estimated by subtracting counterfactual baseline from the social Q-value (see Section \ref{sec-background-ctde}). This advantage enters (\ref{eq:A_tot}) as the social component. We explore two alternative estimates of TD target $y^{SW}$ for COMA that correspond to two definitions of social values, discussed in Section \ref{sec-barocco-sw}. The first estimate (\ref{eq:V_target_prosocial_vanilla}) is based on the common reward and the prediction of COMA for the next state. When $\lambda = 1$, this target is equivalent to the target used in Vanilla COMA. The second estimate (\ref{eq:V_target_prosocial_barocco}), used in BAROCCO, is based on TD targets for the selfish components $y_{i_t}$.


\begin{equation}
\label{eq:V_target_prosocial_vanilla}
  y_{i_t}^{SW_S}=SW(\textbf{r}_t) + \gamma Q_i^{SW}(s_{i_{t+1}}, a_{i_{t+1}})
\end{equation}

\begin{equation}
\label{eq:V_target_prosocial_barocco}
y_t^{SW_L}=SW(\textbf{r}_t + \gamma \textbf{V}(s_{t+1})) = SW(\textbf{y}_t)
\end{equation}

In our implementation, neither critics nor policies share weights.

\begin{figure*}[t]
  \centering
  
    \centering
    \begin{subfigure}{.35\textwidth}
      \centering
      \includegraphics[width=\linewidth]{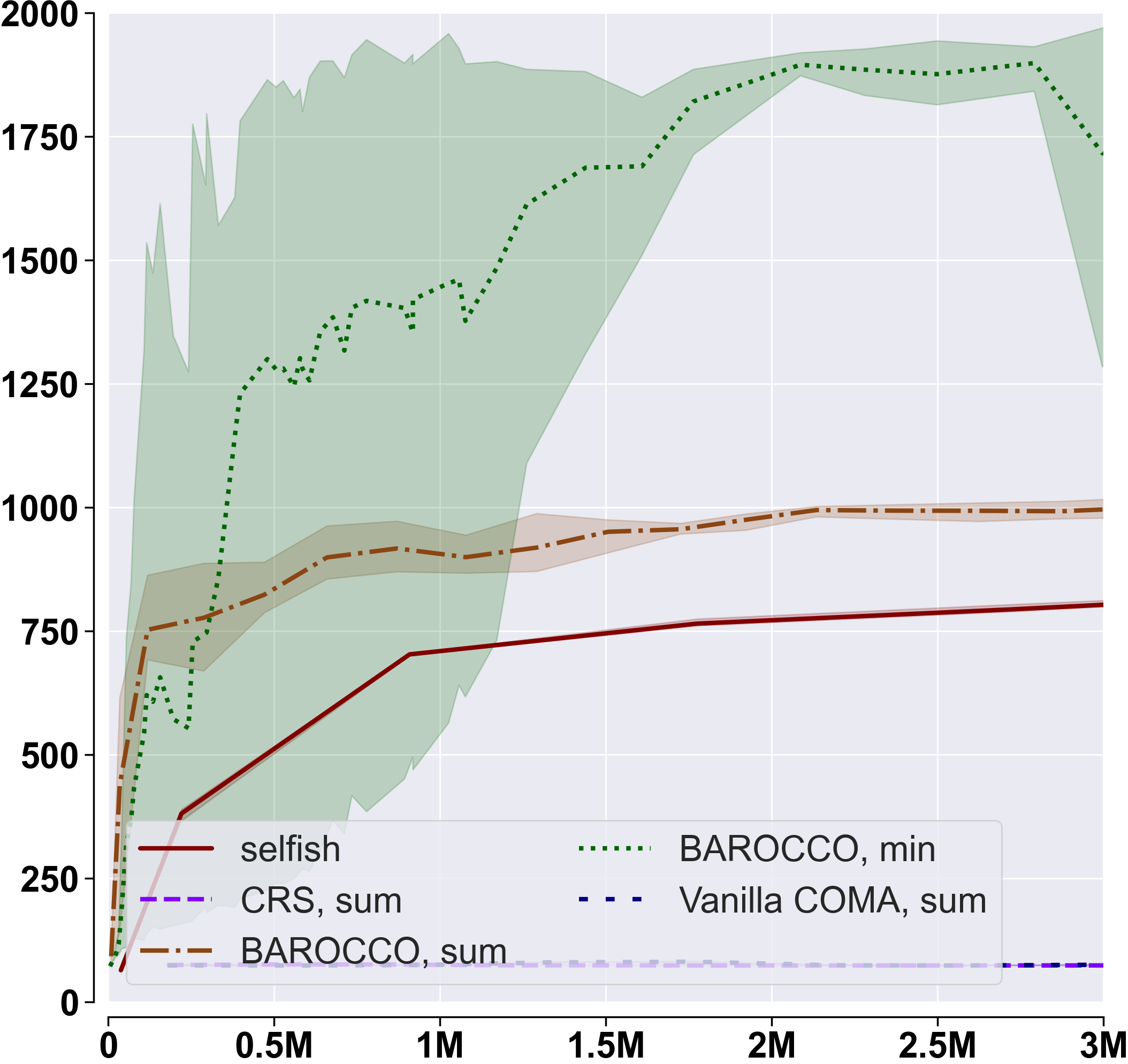}
      \caption{Lifetime, all algorithms}
    \end{subfigure}%
        \begin{subfigure}{.35\textwidth}
      \centering
      \includegraphics[width=\linewidth]{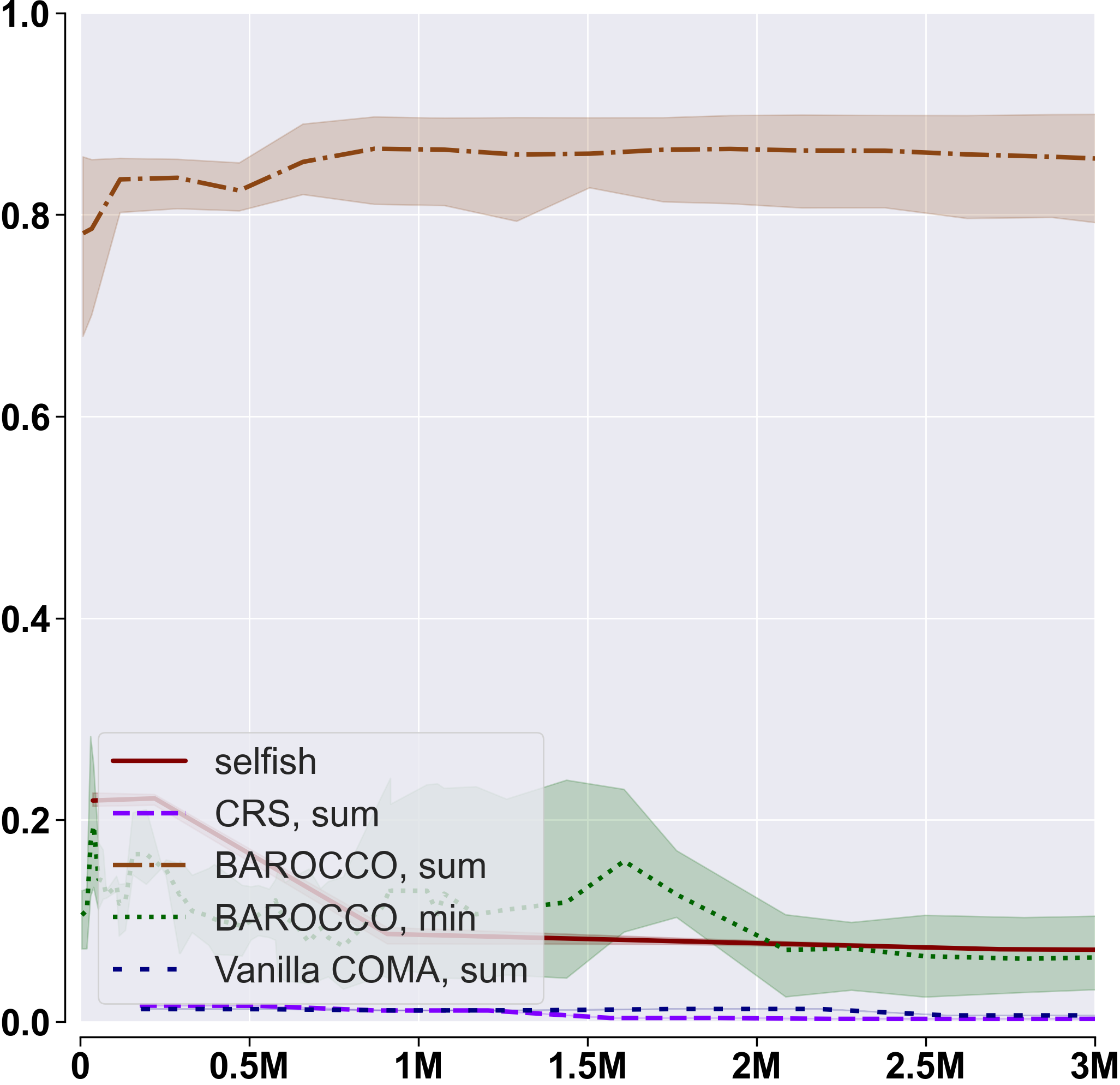}
      \caption{Gini, all algorithms}
    \end{subfigure}
    
    \centering
        \begin{subfigure}{.35\textwidth}
      \centering
      \includegraphics[width=\linewidth]{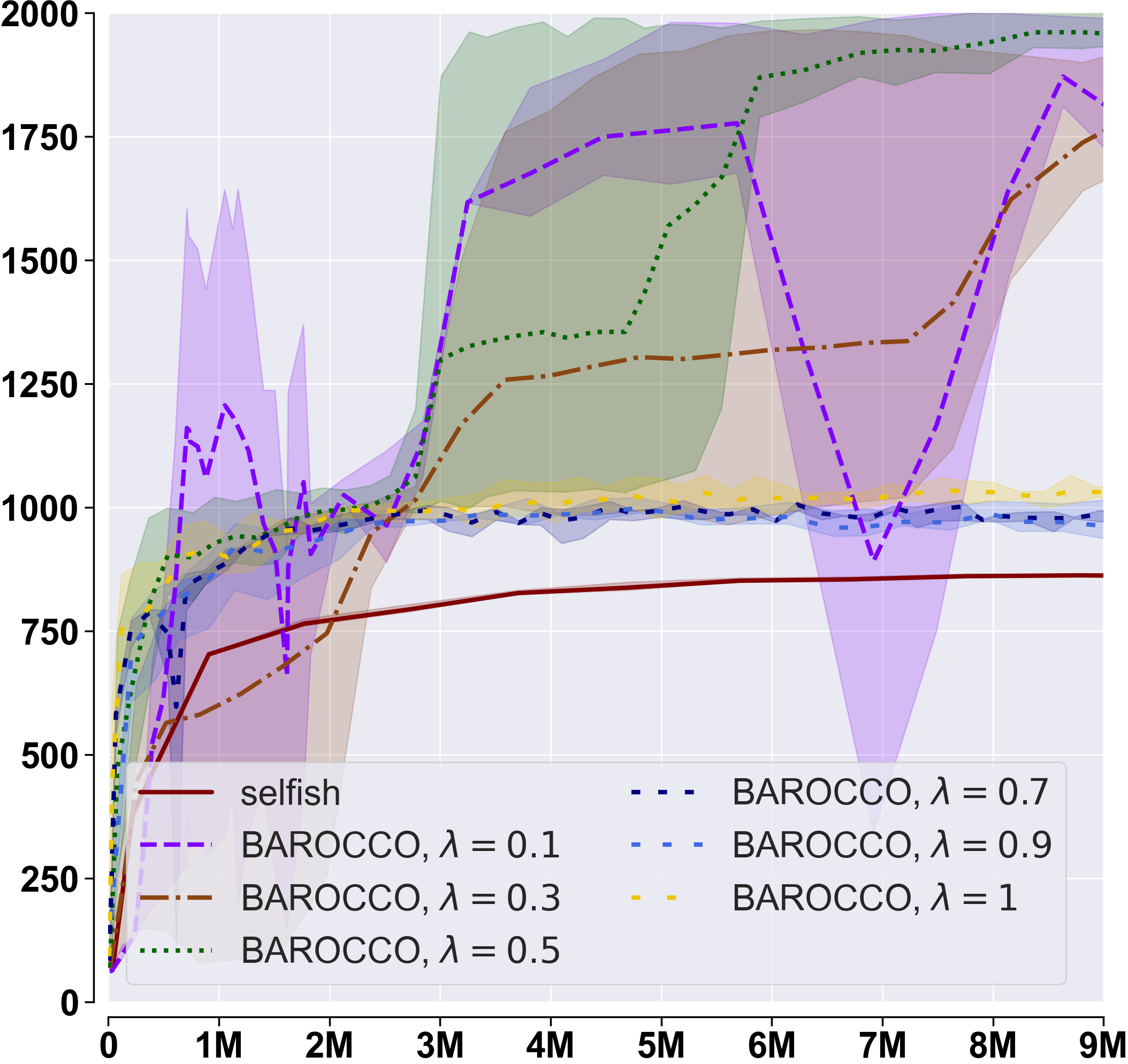}
      \caption{Lifetime, BAROCCO with varying $\lambda$}
    \end{subfigure}%
        \begin{subfigure}{.35\textwidth}
      \centering
      \includegraphics[width=\linewidth]{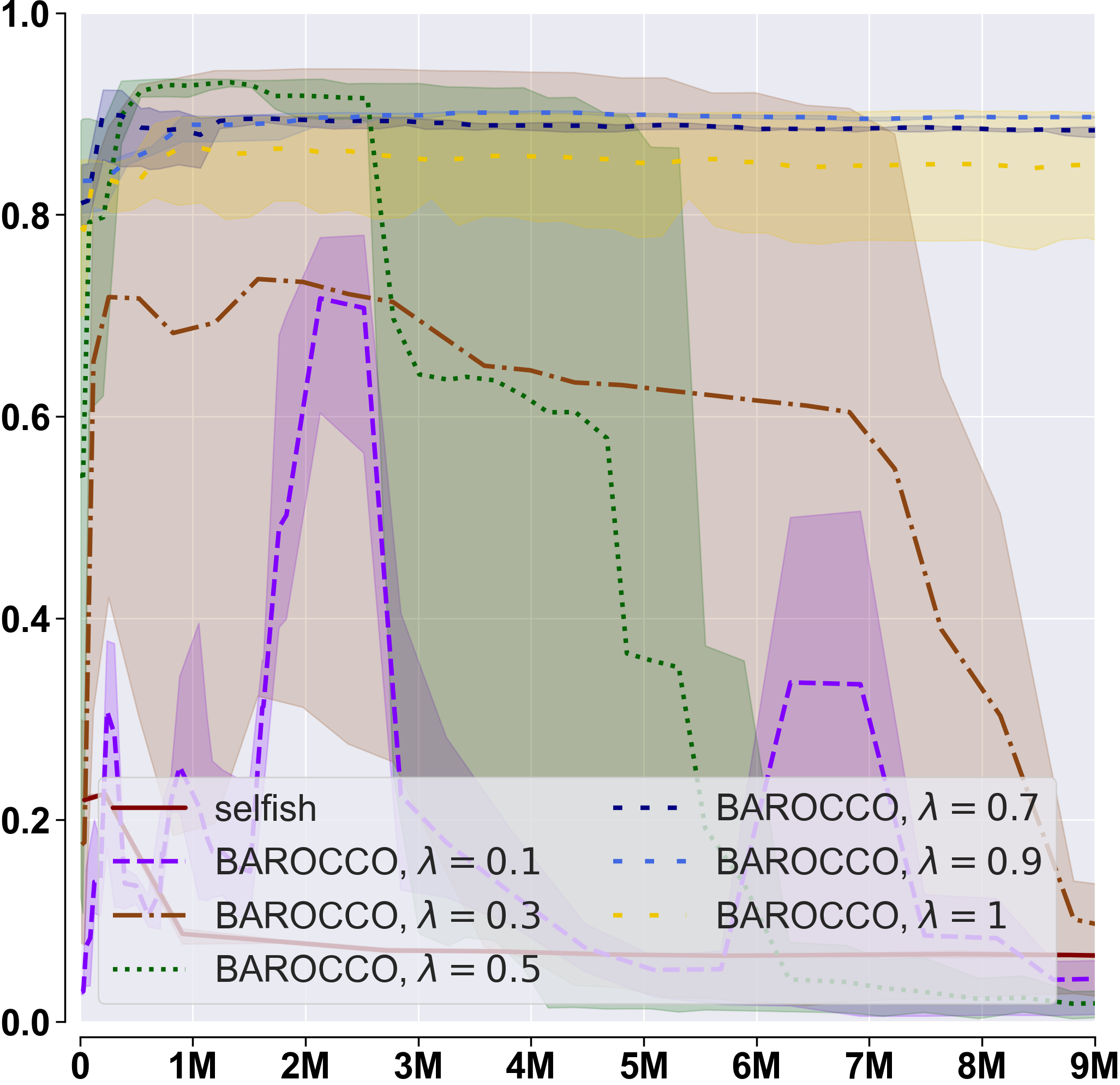}
      \caption{Gini, BAROCCO with varying $\lambda$}
    \end{subfigure}
    
  \caption{Experiments in Eldorado, Actor-Critic framework. `Lifetime` denotes sum of agents' episode lengths, `Gini` is a metric of unfairness. `sum` and `min` denote the choice of $SW$ function.}
  \label{fig:eldorado_ppo}
\end{figure*}

\begin{figure*}[t]
  \centering
  
    \centering
    \begin{subfigure}{.35\textwidth}
      \centering
      \includegraphics[width=\linewidth]{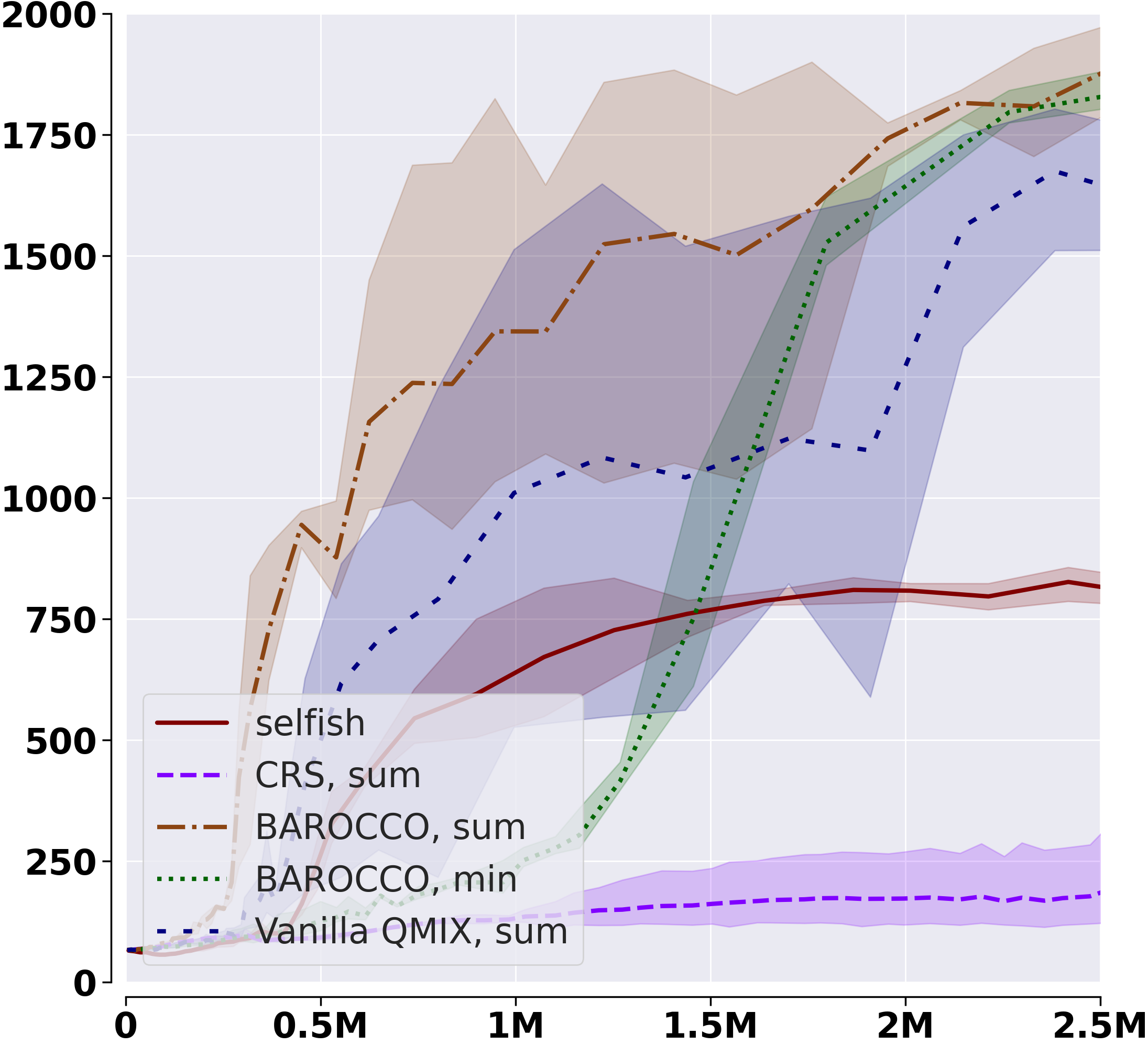}
      \caption{Lifetime, all algorithms}
    \end{subfigure}%
        \begin{subfigure}{.35\textwidth}
      \centering
      \includegraphics[width=\linewidth]{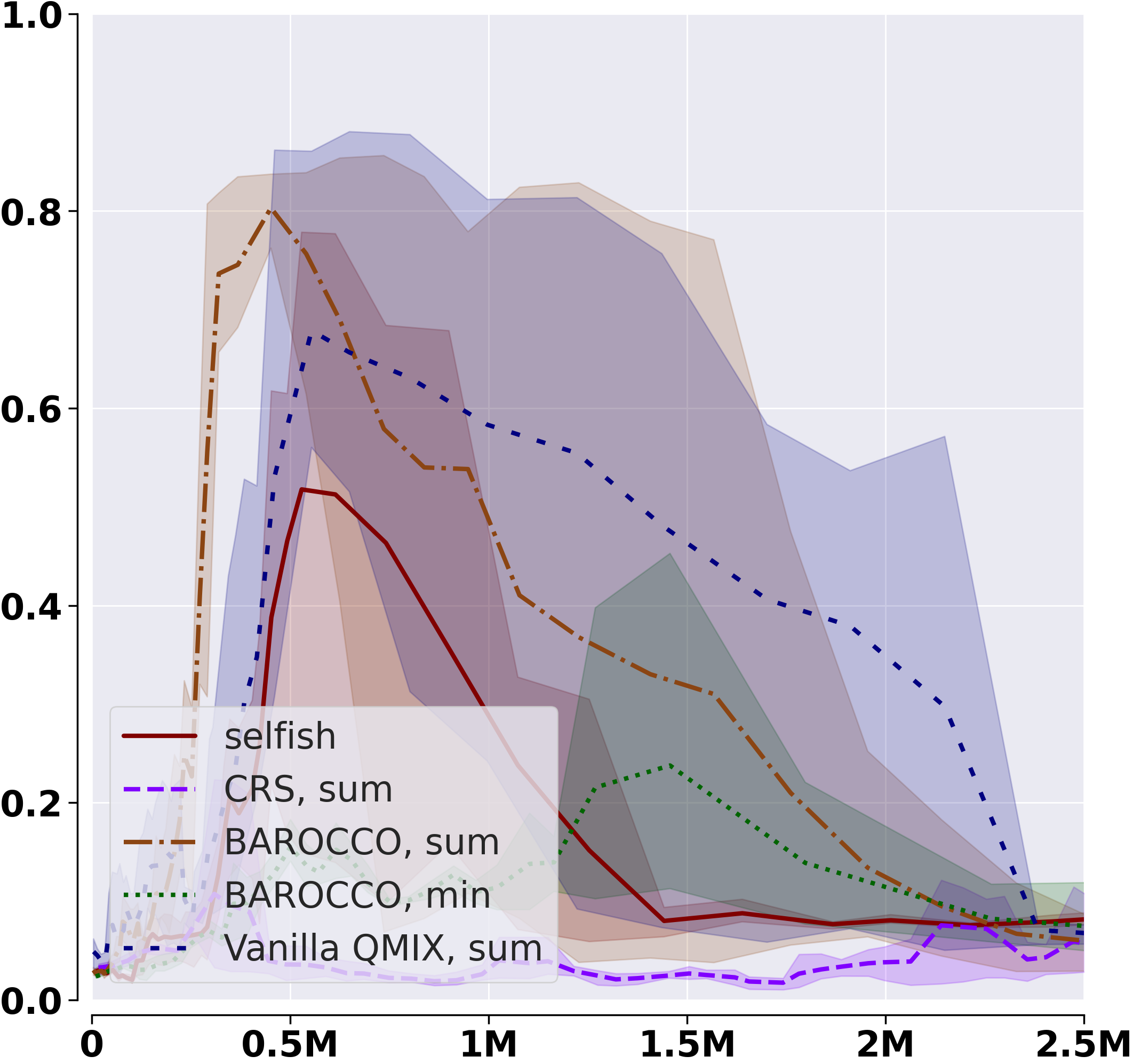}
      \caption{Gini, all algorithms}
    \end{subfigure}
    
    \centering
        \begin{subfigure}{.35\textwidth}
      \centering
      \includegraphics[width=\linewidth]{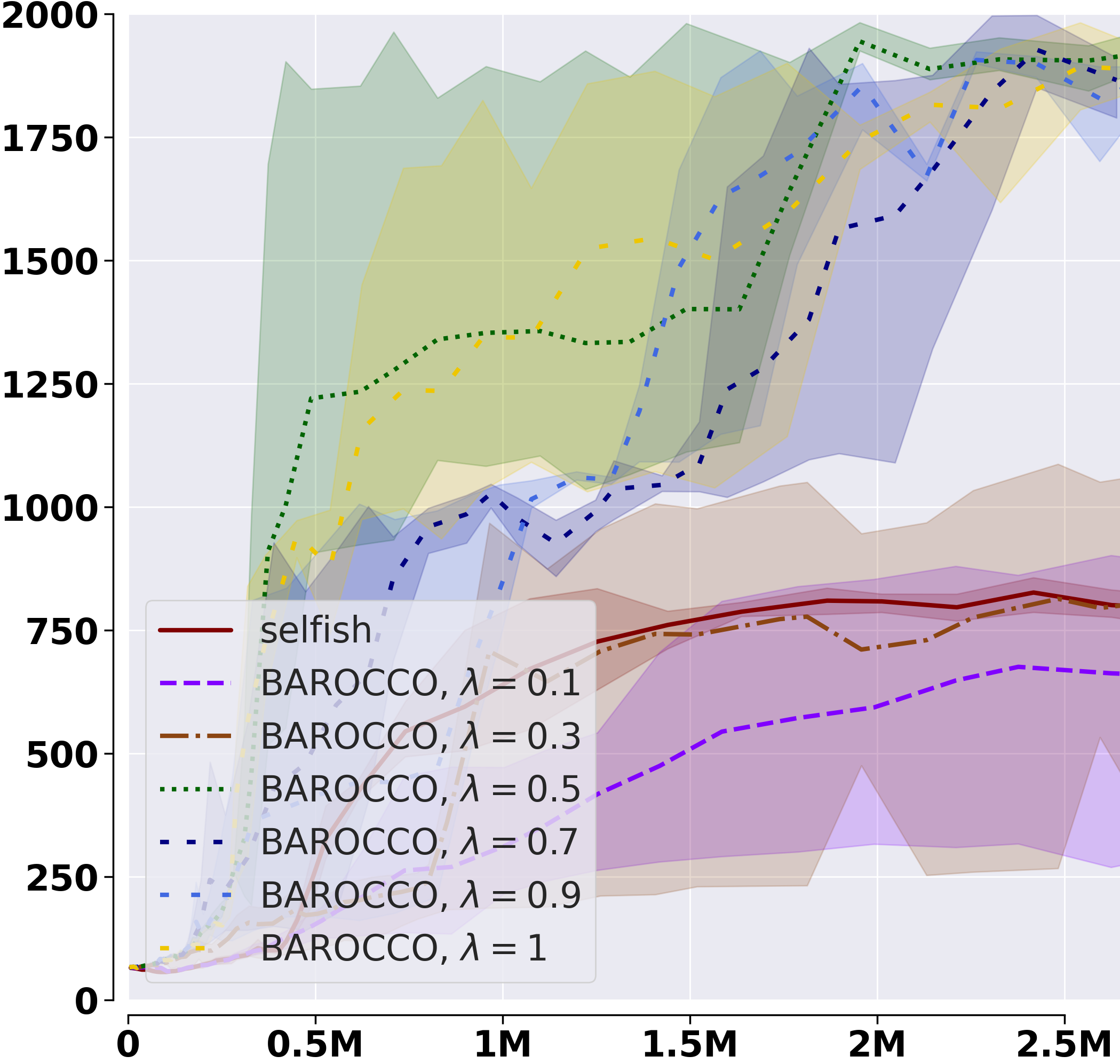}
      \caption{Lifetime, BAROCCO with varying $\lambda$}
    \end{subfigure}%
        \begin{subfigure}{.35\textwidth}
      \centering
      \includegraphics[width=\linewidth]{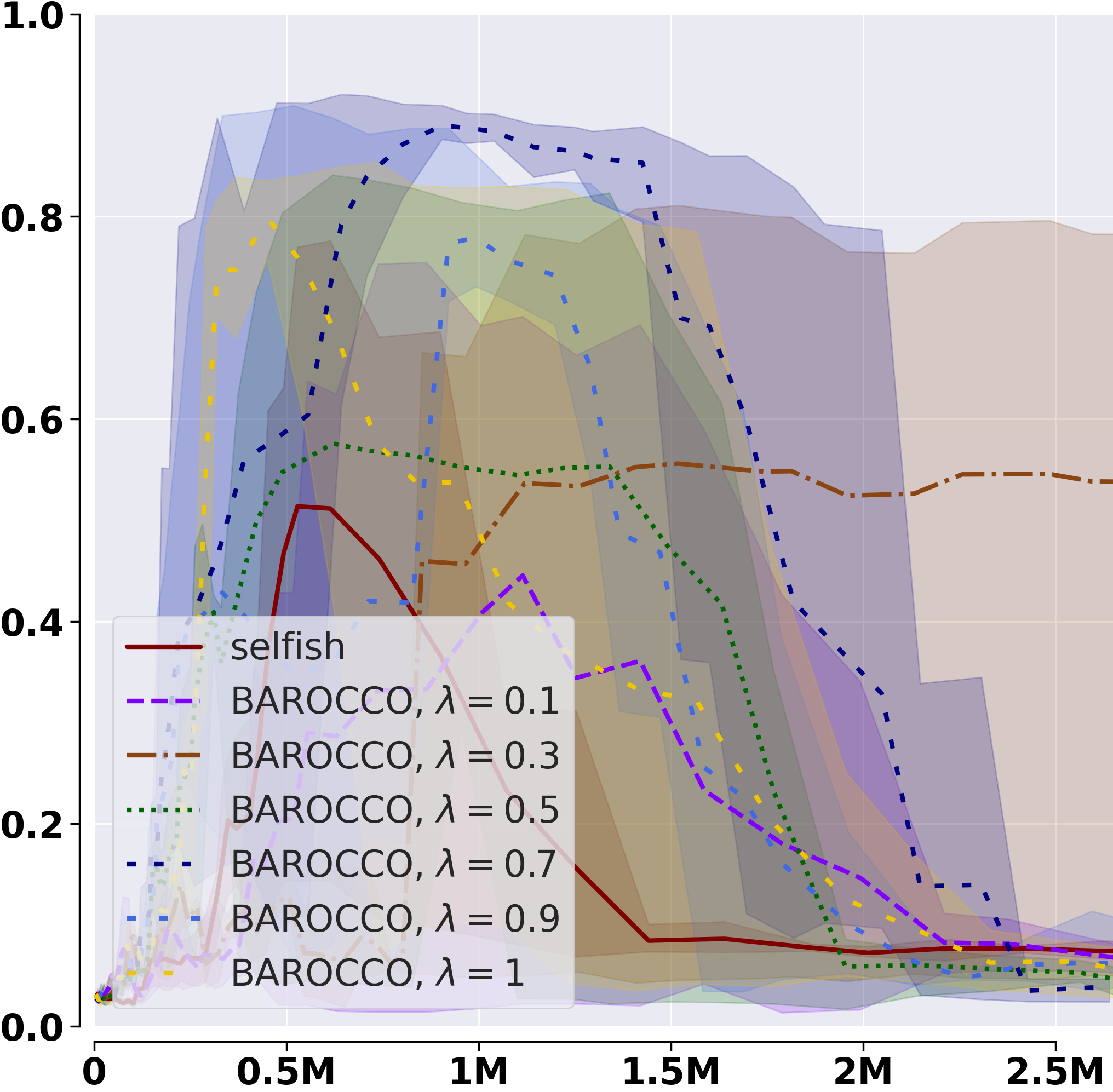}
      \caption{Gini, BAROCCO with varying $\lambda$}
    \end{subfigure}
    
  \caption{Experiments in Eldorado, Q-learning framework. `Lifetime` denotes sum of agents' episode lengths, `Gini` is a metric of unfairness. `sum` and `min` denote the choice of $SW$ function.}
  \label{fig:eldorado_dqn}
\end{figure*}

\section{Experiments}\label{sec-experiments}

\subsection{Modified Prisoner's Dilemma}\label{sec-experiments-mpd}

As a motivational example that illustrates importance of balance between selfish and social incentives, we present modified prisoner's dilemma (Table \ref{table-example}). In this 2 by 3 matrix game, both agents have access to `Cooperate` and `Defect` actions, but one of the agents can also `Sacrifice` his payoffs for the common good. As in the classic prisoner's dilemma \cite{rapoport1965prisoner}, defection is a dominant strategy for a selfish agent. At the same time, mutual defection is Pareto dominated by mutual cooperation. As a result, selfish agents are stuck with mutual defection, even though both agents would benefit from mutual cooperation. In contrast, a social agent prefers to `Cooperate` than to `Defect`.

\begin{table}[h]
\centering
\caption{Modified Prisoner's Dilemma}
\label{table-example}
\begin{tabular}{|c|c|c|c|}
\hline
          & Defect & Cooperate & Sacrifice \\ \hline
Defect    & 5, 5 & 15, 0     & 21, 0   \\ \hline
Cooperate & 0, 15  & 10, 10    & 21, 0     \\ \hline
\end{tabular}
\end{table}

Now, consider the `Sacrifice` action of the column agent. While this action achieves the highest social welfare, it also ensures the worst individual payoff for the second agent. Nevertheless, a social agent always prefers `Sacrifice`, regardless of how small the surplus of social welfare over the mutual cooperation is. Instead, an agent that is willing to cooperate but refuses to self-sacrifice might be preferable.



\begin{table}[h!]
\centering
\caption{Actions of Agents in Modified Prisoner's Dilemma}
\label{table-mpd-actions}
\begin{tabular}{cccc}
\toprule
$\lambda$  & 0-0.3 & 0.4-0.8 & 0.9-1 \\ \midrule
Row player action & D     & C       & C     \\ 
Column player action & D     & C       & S   \\ 
\bottomrule
\end{tabular}
\end{table}

We report behaviour of agents trained to solve Modified Prisoner's Dilemma with tabular Q-learning in Table \ref{table-mpd-actions}. We vary $\lambda$ in $[0, 1]$, each time incrementing it by 0.1. Both agents Defect when $\lambda$ is low and start to Cooperate when $\lambda$ is as high as 0.4. The column agent further switches to Sacrifice when $\lambda$ reaches 0.9. As we will see later in the paper, such sacrificial behaviour is not unique to simple matrix games.

The agents were trained with tabular Q-learning for 100000 iterations. The learning rate was set to 0.1. The exploration rate $\epsilon$ was initialized at 1 and annealed to 0 over the course of training.

\subsection{Environments}\label{sec-environments}


\paragraph{Eldorado.} 

Eldorado (Fig. \ref{fig:map}a) is based on the NeuralMMO environment \cite{suarez2019neural}. Two agents navigate on a fully observable grid-like map, collecting two types of resources -- water and food. Both water and food tiles provide 6 points of the corresponding resource. The water tile has infinite supply, while the food tile has a recharge period of 6 turns. Each agent has limited capacity for the resources, as well as health pool limited to 10 points. Furthermore, both food and water supplies decrease each turn by 1. If some supply is absent, the health points also start to decrease. Conversely, the health regenerates when both supplies are above the threshold of 16. If an agent's health reaches zero, the episode terminates with a unitary negative reward. However, if an agent successfully survives for a 1000 steps, its episode terminates with a unit of positive reward. Upon termination, an agent immediately respawns. Additionally, the agents can interact by attacking each other. This action has two effects. First, it decreases the health of the target by 1. With the small probability of 1/50, the damage is doubled. Second, it steals a unit of both resources. Attack is thus a very appealing action in the short terms. However, in order to successfully complete the task, the agents are required to coordinate their movement while refraining from combat.

\paragraph{Harvest.}

Harvest (Fig. \ref{fig:map}b) is a popular environment \cite{perolat2017multi,hughes2018inequity,jaques2019social} where five agents collect apples on a partially observable grid-like map. Each episode lasts for a thousand steps. The regrowth rate of apples increases with the number of uncollected apples nearby. Therefore, the agents that harvest every apple in sight quickly exhaust the apple supplies. The optimal strategy for a group of agents is to balance harvesting and cultivating apples.



\begin{figure*}[t]
  \centering

    \centering    
    \begin{subfigure}{.35\textwidth}
      \centering
      \includegraphics[width=\linewidth]{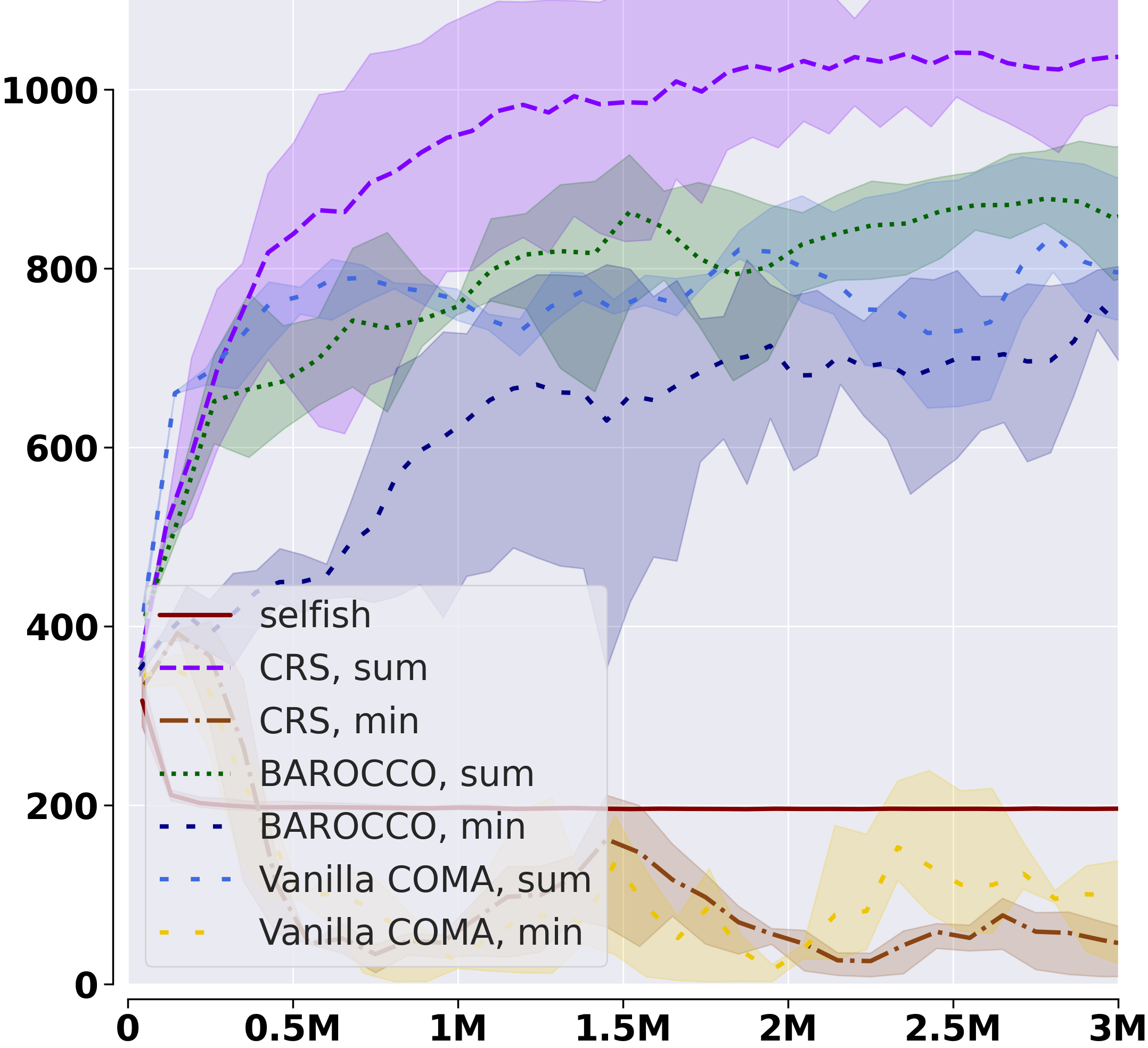}
      \caption{Apples, all algorithms}
    \end{subfigure}%
        \begin{subfigure}{.35\textwidth}
      \centering
      \includegraphics[width=\linewidth]{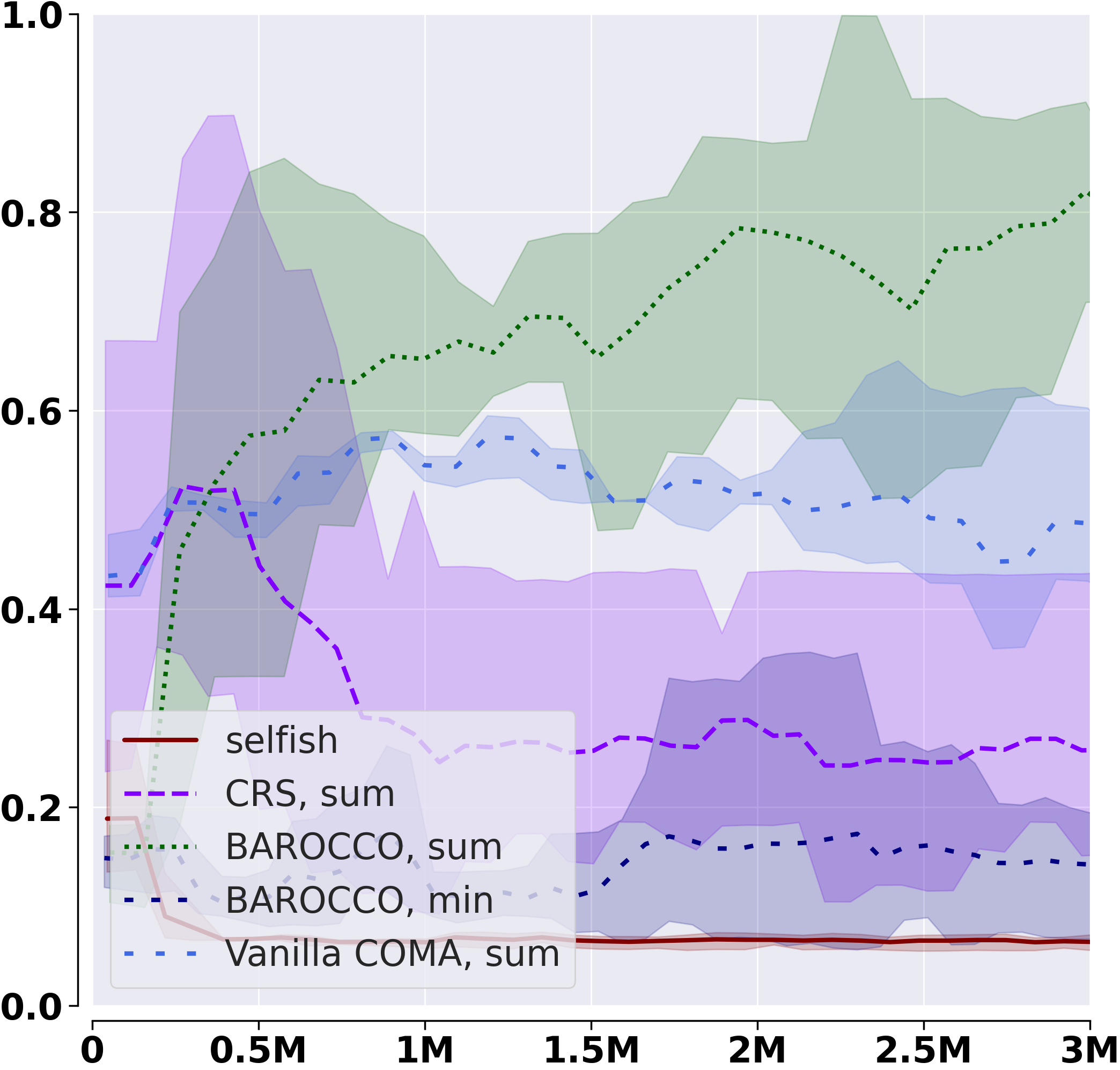}
      \caption{Gini, all algorithms}
    \end{subfigure}

    \centering    
    \begin{subfigure}{.35\textwidth}
      \centering
      \includegraphics[width=\linewidth]{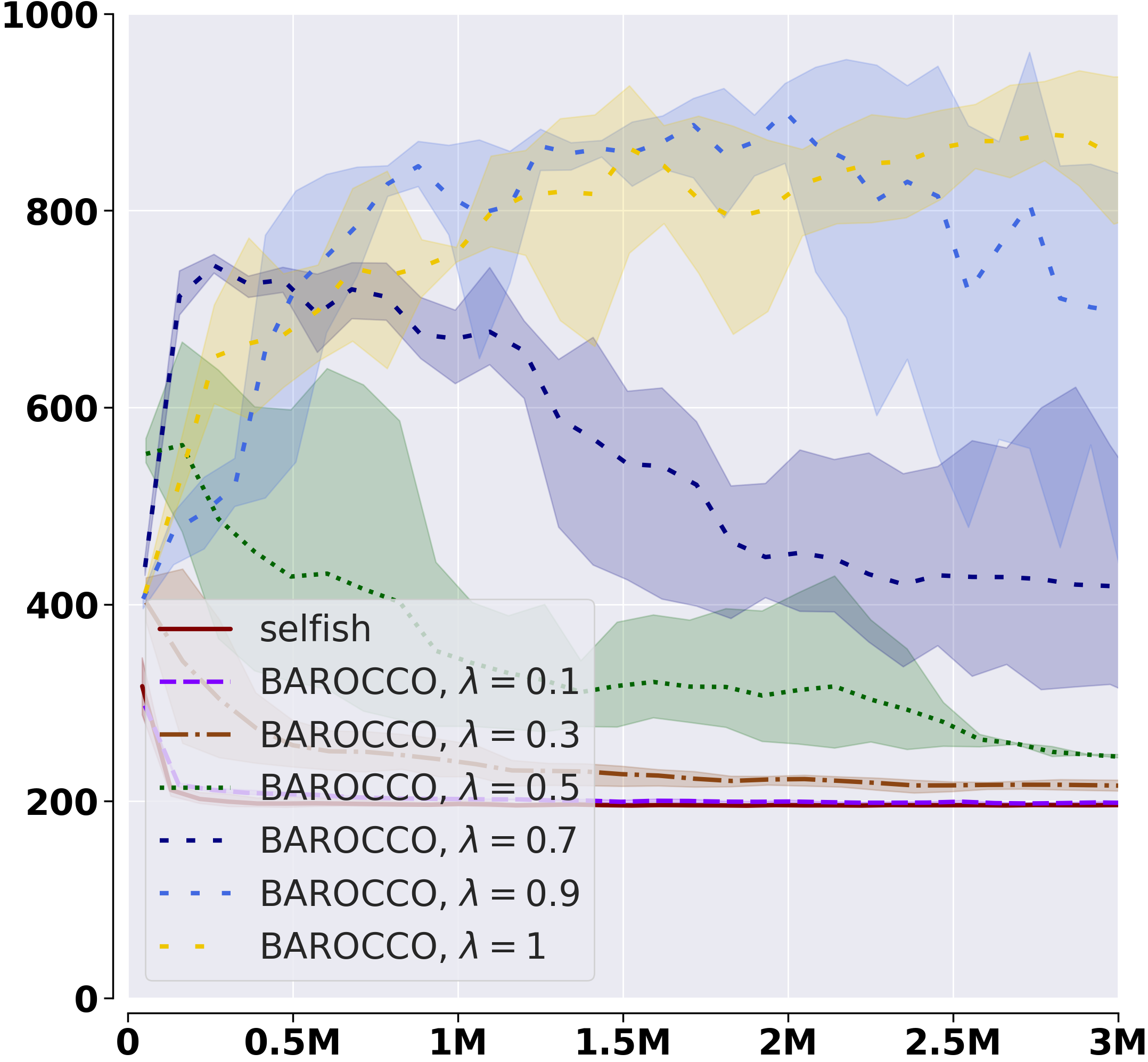}
      \caption{Apples, BAROCCO with varying $\lambda$}
    \end{subfigure}%
        \begin{subfigure}{.35\textwidth}
      \centering
      \includegraphics[width=\linewidth]{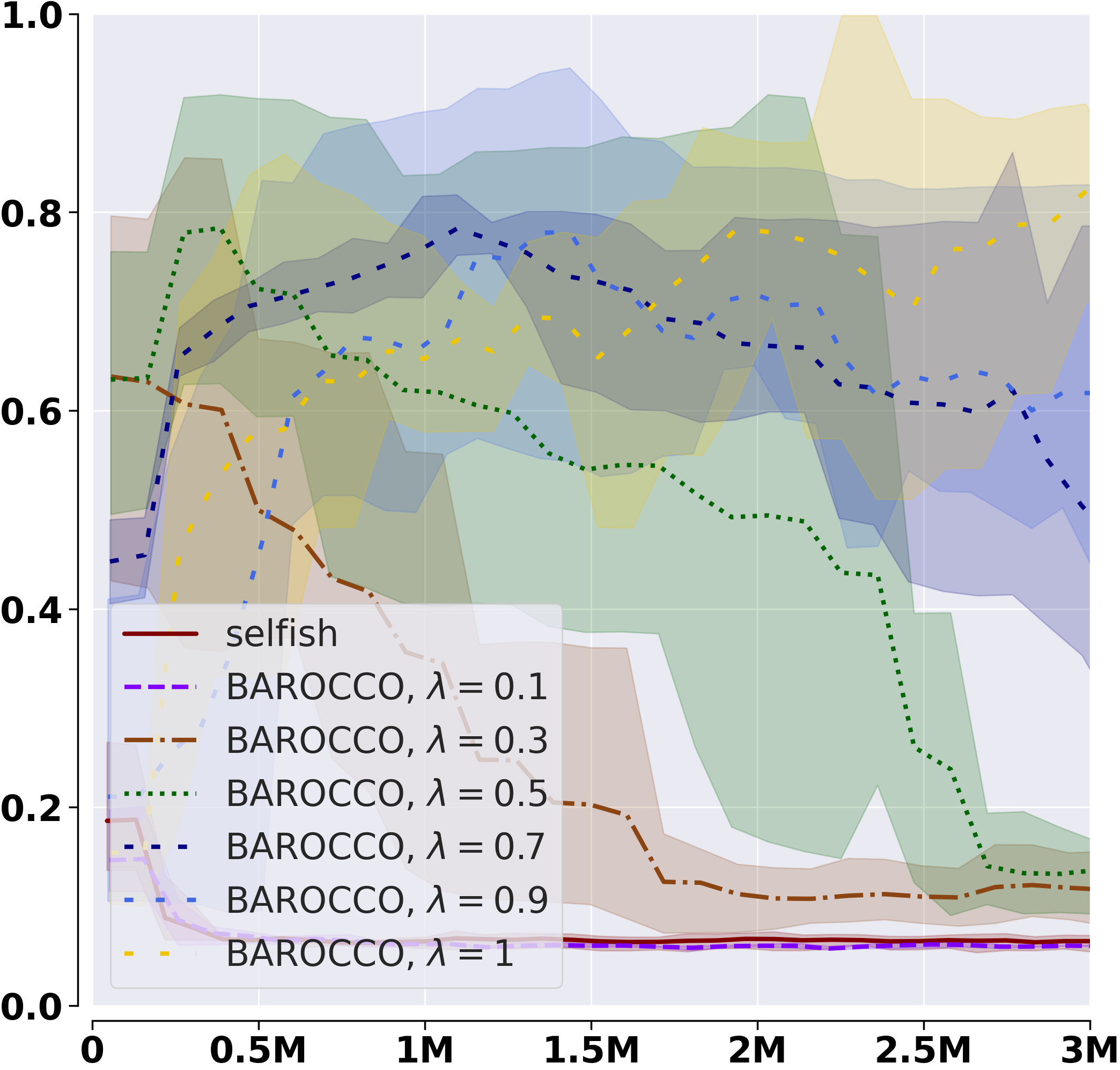}
      \caption{Gini, BAROCCO with varying $\lambda$}
    \end{subfigure}

  \caption{Experiments in Harvest, Actor-Critic framework. `Apples` denotes total number of collected apples by all agents in an episode, `Gini` is a metric of unfairness. `sum` and `min` denote the choice of $SW$ function.}
  \label{fig:harvest_ppo}
\end{figure*}

\begin{figure*}[t]
  \centering
  
      \centering
    \begin{subfigure}{.35\textwidth}
      \centering
      \includegraphics[width=\linewidth]{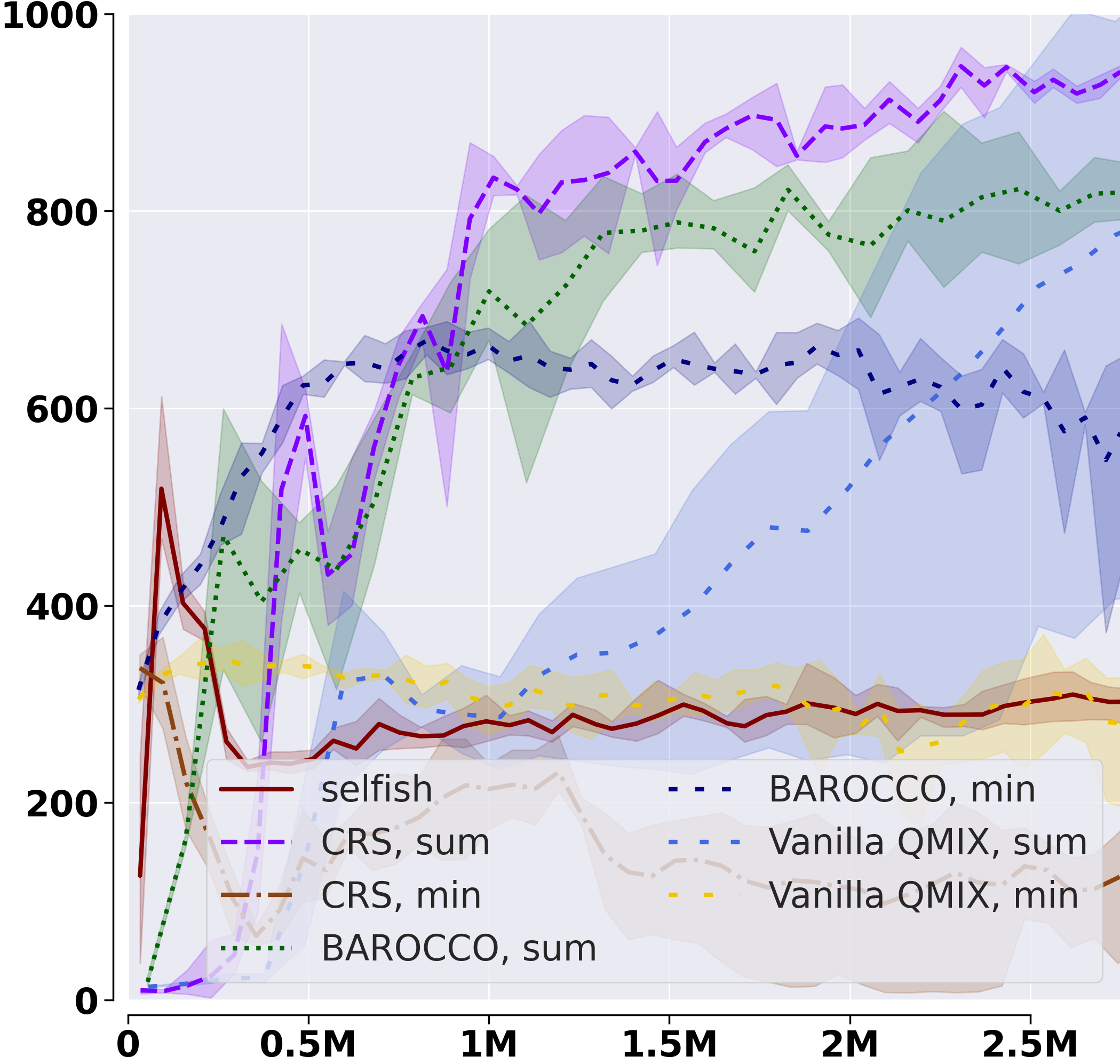}
      \caption{Apples, all algorithms}
    \end{subfigure}%
        \begin{subfigure}{.35\textwidth}
      \centering
      \includegraphics[width=\linewidth]{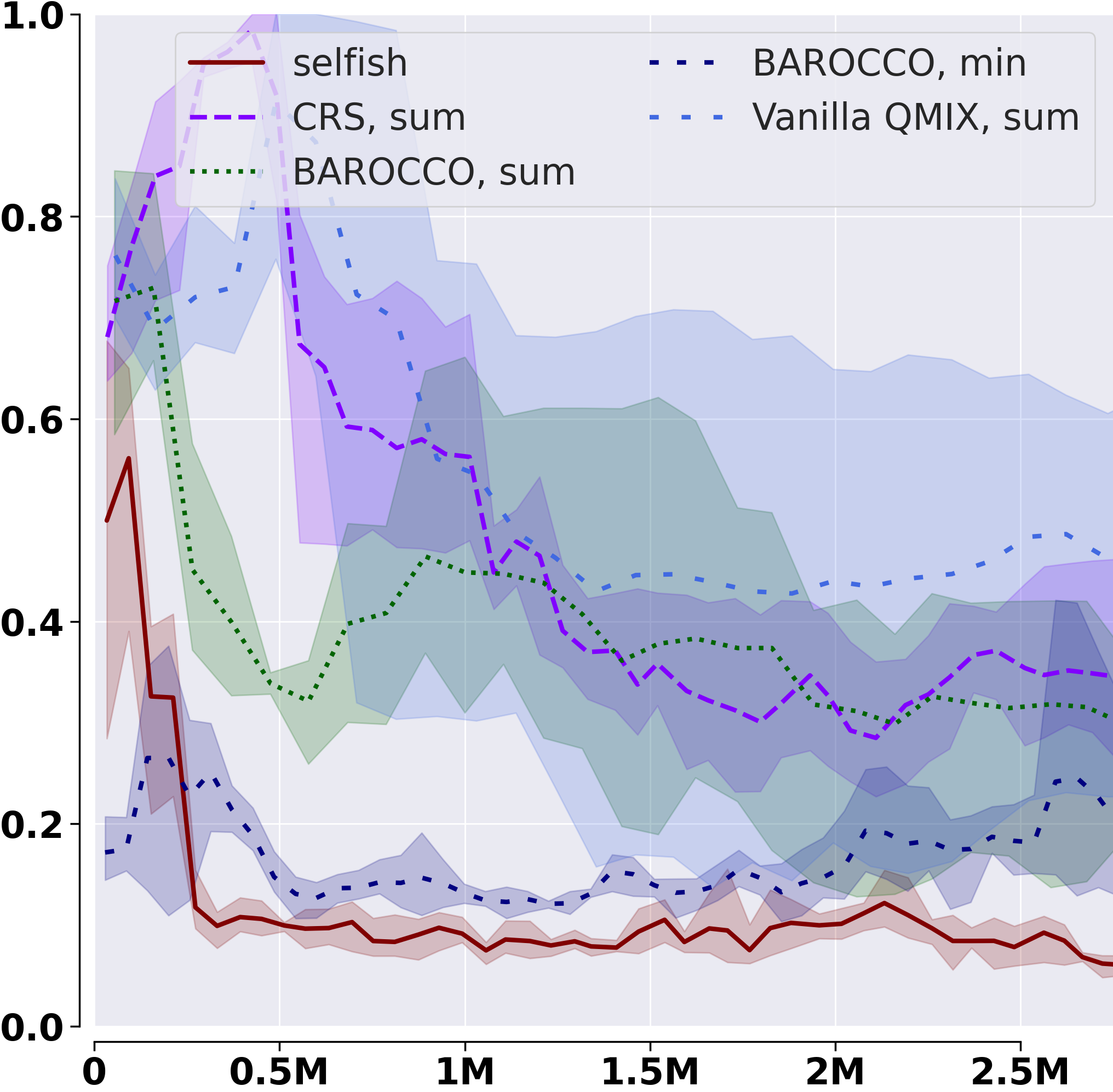}
      \caption{Gini, all algorithms}
    \end{subfigure}
  
    \centering
    \begin{subfigure}{.35\textwidth}
      \centering
      \includegraphics[width=\linewidth]{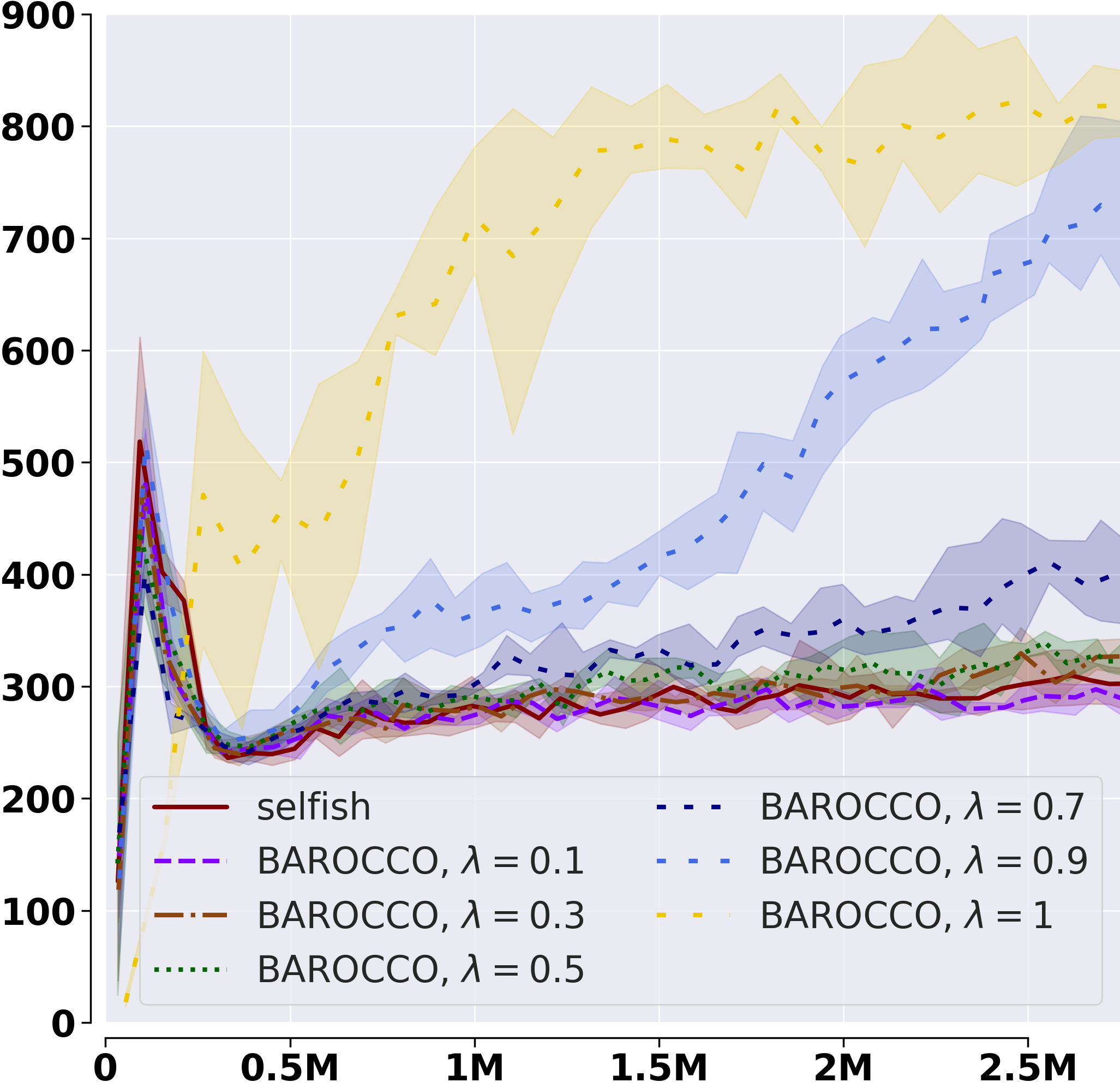}
      \caption{Apples, BAROCCO with varying $\lambda$}
    \end{subfigure}%
        \begin{subfigure}{.35\textwidth}
      \centering
      \includegraphics[width=\linewidth]{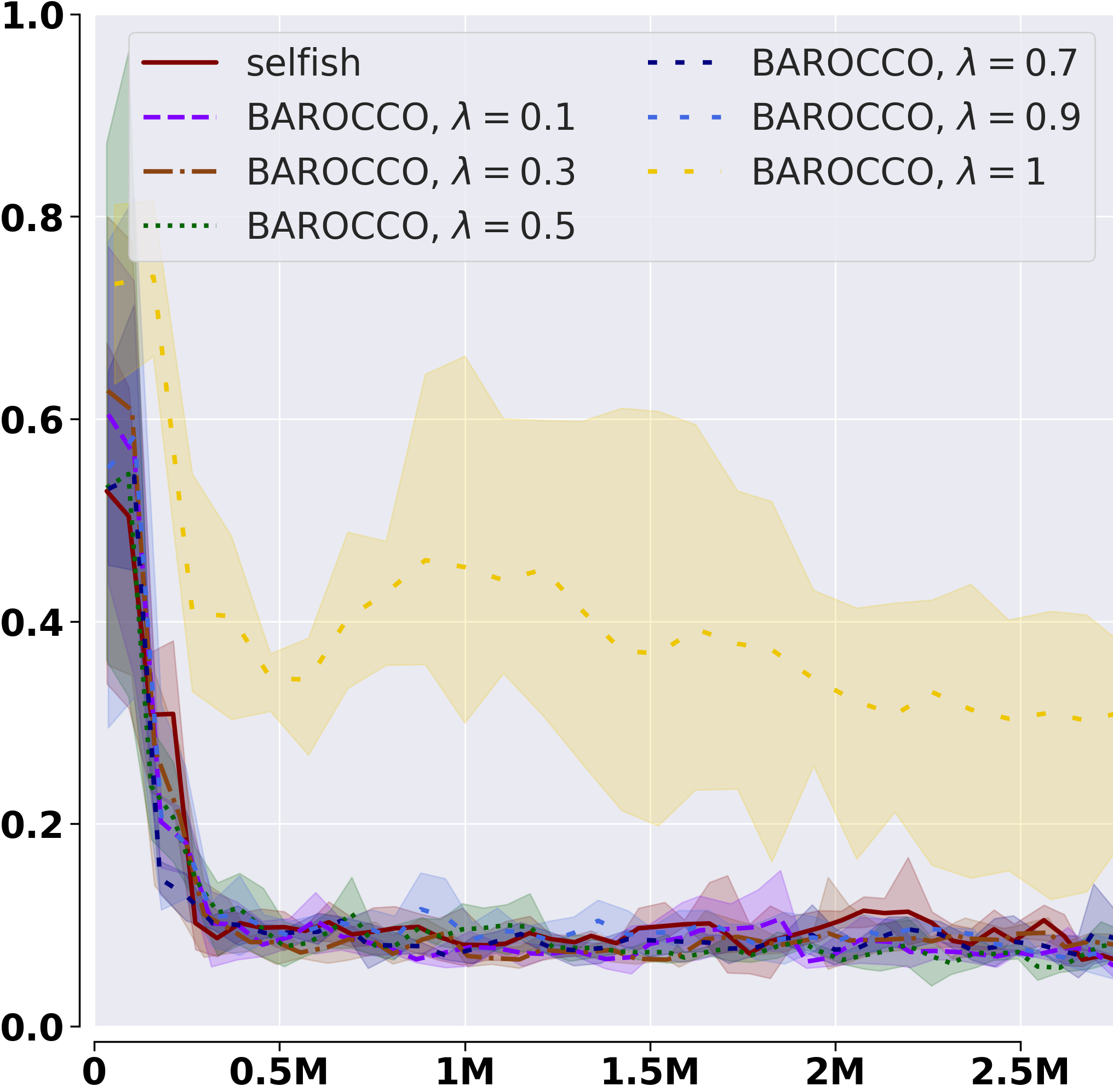}
      \caption{Gini, BAROCCO with varying $\lambda$}
    \end{subfigure}

  \caption{Experiments in Harvest, Q-learning framework. `Apples` denotes total number of collected apples by all agents in an episode, `Gini` is a metric of unfairness. `sum` and `min` denote the choice of $SW$ function.}
  \label{fig:harvest_dqn}
\end{figure*}

\subsection{Results}\label{sec-experiments-results}

We report experimental results for Eldorado and Harvest environments in Figures \ref{fig:eldorado_ppo}, \ref{fig:eldorado_dqn} and \ref{fig:harvest_ppo}, \ref{fig:harvest_dqn}, respectively. We investigate how varying $\lambda$ affects agents' behaviour and performance, as well as compare BAROCCO to baselines, such as selfish baseline, CRS, and Vanilla QMIX / COMA. CRS and Vanilla QMIX / COMA are defined in sections \ref{sec-background-crs} and \ref{sec-barocco-sw}, respectively. Selfish baseline is defined as BAROCCO without the social component, i.e. with $\lambda = 0$. For other algorithms, $\lambda=1$ unless stated otherwise. The algorithms are compared by performance, defined as sum of payoffs, and by fairness, defined according to \cite{perolat2017multi} as unity minus Gini index. We repeat each experiment 3 times. Technical details and hyperparameters are reported in Appendix.

\subsubsection{Actor-Critic agents in Eldorado}

\begin{itemize}
    \item Selfish agents are able to coordinate movement, but are unable to refrain from attacking, since this action is very appealing in short terms. For this reason, they only achieve average lifetime of 800 (Fig. \ref{fig:eldorado_ppo}a).
    \item Unlike selfish agents, prosocial `BAROCCO, sum` agents achieve higher average lifetime (Fig. \ref{fig:eldorado_ppo}a), but most of it is concentrated in a single agent that collects all resources (Fig. \ref{fig:eldorado_ppo}b). This illustrates how maximizing sum of agents' payoffs can result in unfair reward distribution.
    \item 'BAROCCO, min' agents manage to cooperate and successfully solve the environment, reaching average lifetime close to optimal (Fig. \ref{fig:eldorado_ppo}a). This illustrates how optimizing minimum of agents' payoffs instead of sum favours the solutions where payoffs are distributed evenly.
    \item Increasing influence of the selfish component is another way to reject solutions with uneven payoff distribution. When decreasing prosociality coefficient $\lambda$, `BAROCCO, sum` agents are able to escape the local optimum where one of the agents is exploited and learn to both successfully complete the task (Fig. \ref{fig:eldorado_ppo}c,d). This illustrates how fairness emerges from selfishness. The best performance is achieved when $\lambda = 0.5$. It might seem counter-intuitive that the decreasing agents' prosociality positively affects performance, but similar results were reported by \citet{durugkar2020balancing}.
    \item Finally, CRS and COMA agents perform abysmal in Eldorado (Fig. \ref{fig:eldorado_ppo}a). Since these algorithms optimize common reward that is always non-positive in Eldorado, each agent attempts to avoid the negative reward for termination of the other agent and races to terminate earlier, as discussed in Section \ref{sec-barocco-sw}.
\end{itemize}

\subsubsection{Q-learning agents in Eldorado}

\begin{itemize}
    \item Selfish Q-learning agents perform about as good as selfish Actor-Critic agents, reaching average lifetime of 800 that is evenly distributed (Fig. \ref{fig:eldorado_dqn}a, b). These agents are unable to refrain from attacking.
    \item Procosial `BAROCCO, sum` and `BAROCCO, min` agents are able to cooperate and successfully survive in the environment (Fig. \ref{fig:eldorado_dqn}a, b). Unlike the case of Actor-Critic agents, Q-learning `BAROCCO, sum` agents do not converge to a local optimum where one of the agents is exploited.
    \item CRS again underperforms compared to the selfish baseline. QMIX outperforms the selfish baseline but still performs slightly worse than BAROCCO (Fig. \ref{fig:eldorado_dqn}a).
    \item At first, decreasing prosociality coefficient $\lambda$ has negative but slight effect on agents' performance (Fig. \ref{fig:eldorado_dqn}c). While $\lambda > 0.3$, the agents refrain from attacking and manage to survive in the environment. However, once $\lambda$ is at least as low as 0.3, survivability drops significantly as agents begin to combat. The existence of such threshold is consistent with the game-theoretic analysis of \citet{durugkar2020balancing}, as well as with our toy experiment in Section \ref{sec-experiments-mpd}.
\end{itemize}

\subsubsection{Actor-Critic agents in Harvest}

\begin{itemize}
    \item Selfish agents quickly learn to naively harvest every apple in sight and exhaust the supplies long before episode ends. Unable to collude, they only gather about 200 apples per episode (Fig. \ref{fig:harvest_ppo}a).
    \item `BAROCCO, sum` agents learn to alternate between harvesting and cultivating apples and manage to gather more than 800 apples per episode (Fig. \ref{fig:harvest_ppo}a).
    \item `BAROCCO, min` agents collect slightly less apples than `BAROCCO, sum` agents (Fig. \ref{fig:harvest_ppo}a), but distribute the apples significantly more evenly among themselves (Fig. \ref{fig:harvest_ppo}b). This result highlights that optimizing minimum instead of sum of agents' payoffs might be preferable if fairness is a concern.
    \item `COMA, sum` also outperforms selfish agents but is less stable than `BAROCCO, sum` (Fig. \ref{fig:harvest_ppo}a), meaning that our modifications of the training procedure can be beneficial.
    \item Decentralized `CRS, sum` performs better than all other algorithms (Fig. \ref{fig:harvest_ppo}a), suggesting that additional complexity of centralized algorithms can hinder performance in some environments. This result contradicts the findings of the prior literature where centralization of training consistently improved performance \cite{qmix,coma}. However, the algorithms suggested in this literature were not tested in complex mixed environments like Harvest before.
    \item Performance of CRS and COMA plummets when minimum is chosen as $SW$ (Fig. \ref{fig:harvest_ppo}a), which is consistent with our predictions formulated in Section \ref{sec-barocco-sw} that optimizing minimum of agents rewards each step might be too restricting. This result also highlights flexibility of BAROCCO in the choice of $SW$ function.
    \item The effect of varying $\lambda$ is monotonic: increasing $\lambda$ improves performance (Fig. \ref{fig:harvest_ppo}c) but can result in unfair reward allocation (Fig. \ref{fig:harvest_ppo}d).
\end{itemize}

\subsubsection{Q-learning agents in Harvest}

\begin{itemize}
    \item By and large, the results (Fig. \ref{fig:harvest_dqn}) are similar to the case of Actor-Critic agents. Selfish agents converge to a naive strategy of collecting every apple in sight. `BAROCCO, sum` agents outperform selfish agents by balancing harvesting and cultivating apples. `BAROCCO, min` performs a little worse than `BAROCCO, sum` but leads to a more even apple distribution. `QMIX, sum` is less stable than `BAROCCO, sum`, highlighting that our training procedure is more suitable for mixed environments. `CRS, sum` performs better than all other algorithms. `CRS, min` and `QMIX, min` fail to outperform even selfish agents, in contrast to BAROCCO that is flexible in the choice of $SW$ function.
    \item While the best team performance is achieved when prosociality coefficient $\lambda$ is maximal (Fig. \ref{fig:harvest_dqn}c), this solution favours unfair apple allocation (Fig. \ref{fig:harvest_dqn}d). Setting $\lambda=0.9$ leads to slower convergence and slightly lower final team performance, but is a considerably less unfair solution.
\end{itemize}

\section{Conclusion}

%

In this paper, we present BAROCCO -- a meta-algorithm for combining social and selfish incentives in cooperative-competitive environments. We confirm the effectiveness of BAROCCO over the existing methods in two mixed multi-agent environments for both Q-learning and Actor-Critic frameworks. Specifically, we find that BAROCCO consistently improves over vanilla QMIX and COMA in all experiments, highlighting usefulness of the modifications that we propose for training these algorithms in mixed environments. Furthermore, we find that varying the prosociality coefficient $\lambda$ results in unique mixtures of selfish and selfless behaviour. While decreasing $\lambda$ typically increases fairness at the expense of efficiency, in some cases both efficiency and fairness can benefit from the influence of the selfish component. As an alternative way to achieve fairness, BAROCCO also allows to train fair cooperative agents by maximizing minimum of selfish payoffs. An exciting extension of our work could be to train reciprocal agents that dynamically assess the cooperativeness of others and adapt their policies accordingly. We also note that BAROCCO is not limited to the algorithms utilized in this paper, i.e. DQN, PPO, MADDPG, COMA and QMIX. Rather, we propose a unified framework of two separate modules, which can be modified by other state-of-the-art techniques from single-agent, mixed, or cooperative setups.

This work contributes to the broader discussion of what constitutes cooperation. Most MARL papers that study mixed environments focus on efficiency, but we argue that this metric can be too limiting. Agents that act towards a single common goal are more reminiscent of a swarm system than a group of distinct individuals that could mutually benefit from cooperation. We explore ways to incorporate the notion of fairness into such systems, either by preserving some individuality of the agents or by modifying the centralized objective. We hope that our work sparks further discussion regarding other desirable qualities of multi-agent systems and the means to achieve these qualities.




\section{Acknowledgements}

This research was supported in part through computational resources of HPC facilities at HSE University. Support from the Basic Research Program of the National Research University Higher School of Economics is gratefully acknowledged.


\appendix

\section{Assessing Social Welfare: Examples}

In this section, we elaborate on the advantages of the long-term value approach formulated in Section \ref{sec-barocco-sw}.

\paragraph{Applicability of sum and minimum as $SW$.}

Consider the following toy environment. A centralized controller distributes positive unitary rewards between two agents for two time-steps. Furthermore, if the same agent is rewarded twice, the second reward is doubled. In this environment, there are 4 options to distribute rewards: 2 options to reward the same agent at both time-steps, and 2 options to reward one agent at the first time-step and the other agent at the second time-step. We are interested how to distribute the rewards in order to maximize social welfare. We analyze two definitions of the prosocial value function, i.e. $V^{SW_S}$ and $V^{SW_L}$, as well as two choices of $SW$ function, i.e. sum and minimum. The results of the analysis are summarized in Table \ref{tab:example_min}.

\begin{table*}[h]
    \centering
    \caption{Reward distributions and corresponding social welfare in the toy environment}
\begin{tabular}{@{}ccccccc@{}}
\toprule
         & \multicolumn{2}{c}{}                    & \multicolumn{2}{c}{$SW$ is sum} & \multicolumn{2}{c}{$SW$ is min} \\
         & Rewards of agent 1 & Rewards of agent 2 & $V^{SW_S}$     & $V^{SW_L}$     & $V^{SW_S}$     & $V^{SW_L}$     \\ \midrule
Option 1 & $[1, 2\gamma]$     & $[0, 0]$           & $1 + 2\gamma$  & $1 + 2\gamma$  & 0              & 0              \\
Option 2 & $[1, 0]$           & $[0, \gamma]$      & $1 + \gamma$   & $1 + \gamma$   & 0              & $\gamma$       \\
Option 3 & $[0, \gamma]$      & $[1, 0]$           & $1 + \gamma$   & $1 + \gamma$   & 0              & $\gamma$       \\
Option 4 & $[0, 0]$           & $[1, 2\gamma]$     & $1 + 2\gamma$  & $1 + 2\gamma$  & 0              & 0              \\ \bottomrule
\end{tabular}
\label{tab:example_min}
\end{table*}

We can observe several patterns consistent with our experimental findings. First, when $SW$ is chosen as sum, the value functions $V^{SW_S}$ and $V^{SW_L}$ are equivalent. This is a consequence of commutativity of sum with itself: the order of summation over time-steps and over agents does not affect on resulting value function. Furthermore, maximization of the social welfare requires to sacrifice the interests of one of the agents by choosing either option 1 or 4. Second, when $SW$ is chosen as minimum, all four options are equivalent from the standpoint of $V^{SW_S}$. This is a consequence of the environment design: the controller is unable to reward both agents at the same time-step and minimum of two rewards is always 0. In contrast, maximization of $V^{SW_L}$ requires fair reward distribution on average, and thus options 2 and 3 are preferred. Therefore, if fairness is a concern, $SW$ should be chosen as minimum and $V^{SW_L}$, i.e. the long-term approach to define value function that is used in BAROCCO,  should be focused on.

\paragraph{Environments where trajectory lengths vary.}

Consider a two-agent environment where the only reward that each agent receives is a unitary negative reward upon termination. We are interested in the incentives that drive selfish and prosocial agents in case of such reward structure. In this example, $SW$ function will be chosen as sum. Let the trajectory lengths of the two agents be $T_1$ and $T_2$, respectively, and let $T_1 < T_2$. The agent that terminates earlier will be referred to as the first agent, and vice versa. The selfish values $V_i$ and the prosocial values $V_i^{SW_S}$ and $V_i^{SW_L}$ of the two agents are estimated in Table \ref{tab:example_env} (expectation operator is omitted).

\begin{table*}[h]
    \centering
    \caption{Values of two agents in an environment with $-1$ reward upon termination ($T_1 < T_2$)}
\begin{tabular}{@{}cccc@{}}
\toprule
                 & Selfish agents                                           & \multicolumn{2}{c}{Prosocial agents}                                                                                                                                                          \\
General formula  & $V_i = \overset{T_i}{\underset{t=0}{\sum}} \gamma^t r_i$ & $V_i^{SW_S} = \overset{T_i}{\underset{t=0}{\sum}} \gamma^t (r_1 + r_2)$ & $V_i^{SW_L} = \overset{T_1}{\underset{t=0}{\sum}}  \gamma^t r_1 + \overset{T_2}{\underset{t=0}{\sum}} \gamma^t r_2$ \\ \midrule
Value of agent 1 & $ -\gamma^{T_1}$                                         & $ -\gamma^{T_1}$                                                        & $ -(\gamma^{T_1} + \gamma^{T_2})$                                                                                   \\
Value of agent 2 & $ -\gamma^{T_2}$                                         & $ -(\gamma^{T_1} + \gamma^{T_2})$                                       & $ -(\gamma^{T_1} + \gamma^{T_2})$                                                                                   \\ \bottomrule
\end{tabular}
    \label{tab:example_env}
\end{table*}

Depending on the value function that the agents optimize, they might learn different behaviour. First, each of the selfish agents is only incentivized to prolong its own trajectory. As was shown in the literature, such agents may struggle to achieve mutual benefits of stable cooperation \cite{peysakhovich2018prosocial,hughes2018inequity,jaques2019social,wang2019evolving}. In contrast, the prosocial agents that optimize $V_i^{SW_L}$ are incentivized to prolong the trajectories of both agents and thus are willing to cooperate. However, this is not the only incentive that drives the prosocial agents that optimize $V_i^{SW_S}$. While both such agents do benefit from longer episodes, each agent also prefers to be the first agent rather than the second, i.e. terminate earlier. This incentive emerges because the first agent does not observe termination of the second agent. Moreover, by comparing values of such agents (Table \ref{tab:example_env}, column 2) it is evident that the first agent receives higher payoffs than the second \textit{regardless} of how long the second agent survives, since $-\gamma^{T_2} \leq 0$. Therefore, instead of cooperating to survive, such agents would compete for early termination.

A similar analysis can be performed for the opposite kind of environments where the termination reward is positive. In such environments, the agents are usually required to complete certain tasks. Instead, the agents that optimize $V^{SW_S}$ would delay task completion in attempts to observe termination of the others. 

The two discussed environments with positive and negative termination rewards are extreme examples with two opposite artifacts. However, a combination of these artifacts may emerge in an environment with an arbitrary reward structure and varying episode length, which can result in unexpected and suboptimal behaviour.

\begin{table*}[h]
\centering

\caption{Hyperparameters}

\begin{tabular}{@{}lcccc@{}}
\toprule
                                 & \multicolumn{2}{c}{Q-learning}        & \multicolumn{2}{c}{Actor-Critic}            \\ 
                                 & Eldorado          & Harvest           & Eldorado             & Harvest              \\ \midrule
discount factor $\gamma$         & 0.99              & 0.99              & 0.99                 & 0.99                 \\
Adam learning rate               & 0.0005            & 0.0005            & 0.0005               & 0.001                \\
learning rate decay              & 0.999995          & 0.999995          & 0.999998             & 0.9998               \\
batch size                       & 64                & 128               & 2000                 & 3000                 \\
mini-batch size                  & \multicolumn{2}{c}{-}                 & 500                  & 500                  \\
\# epochs                        & \multicolumn{2}{c}{-}                 & 10                   & 3                    \\
\# FC layers                     & 3                 & 2                 & 2                    & 3                    \\
\# per-layer FC neurons          & 64                & 64                & 128                  & 64                   \\
\# LSTM layers                   & 0                 & 0                 & 0                    & 1                    \\
\# CNN layers                    & 0                 & 1                 & 0                    & 1                    \\
target network period            & 2K                & 2K                & \multicolumn{2}{c}{-}                       \\
exploration rate $\epsilon$      & $1 \rightarrow 0$ & $1 \rightarrow 0$ & \multicolumn{2}{c}{-}                       \\
$\epsilon$ decay                 & 0.99999           & 0.999975          & \multicolumn{2}{c}{-}                       \\
noisy exploration $\sigma_0$     & 0.5               & 0.5               & \multicolumn{2}{c}{-}                       \\
entropy coefficient $\beta$      & \multicolumn{2}{c}{-}                 & $0.05 \rightarrow 0$ & $0.05 \rightarrow 0$ \\
entropy decay                    & \multicolumn{2}{c}{-}                 & 0.99998              & 0.998                \\
buffer size                      & 500K              & 250K              & \multicolumn{2}{c}{-}                       \\
prioritization exponent          & 0.6               & 0.6               & \multicolumn{2}{c}{-}                       \\
\# quantiles (selfish component) & 10                & 10                & 1                    & 1                    \\
\# steps in $n$-step returns     & 5                 & 5                 & 1                    & 1                    \\ \bottomrule
\end{tabular}
\label{table:hyperparameters}
\end{table*}

\section{Technical Details}

Pseudocode of BAROCCO for Q-learning and Actor-Critic agents is presented in Algorithms \ref{alg:barocco_q} and \ref{alg:barocco_ac}, respectively. The choice of hyperparameters for the algorithms is reported in Table \ref{table:hyperparameters}.

In Q-learning framework, the selfish component is implemented via Rainbow \cite{rainbow}, and the prosocial component is implemented via QMIX \cite{qmix}. Neither of the components utilizes parameter sharing for $Q_i(o_i, a_i)$ or $Q_i^{SW}(o_i, a_i)$ predictions. Both noisy \cite{noise} and $\epsilon$-greedy explorations are applied. The rate of exploration $\epsilon$ is annealed to 0. This is an important detail, because for a given agent the hard-coded randomness of other agents' actions can change its optimal policy \cite{wunder2010classes}. Both Rainbow and QMIX use experience replay buffers. A well-known issue of experience replay is that it can be harmful in non-stationary environments \cite{lin1992self}. To address the inherent non-stationarity of multi-agent environments, we adopt the fingerprint technique \cite{fingerprint} by adding the exploration rate $\epsilon$ to the state space. Vanilla QMIX additionally utilizes double Q-learning \cite{double,fureducing}. Finally, we utilize multiprocessing to perform interaction with environment, update of the selfish components, and update of the prosocial components in parallel, similarly to APEX \cite{apex}.

In Actor-Critic framework, the selfish component for each agent is a critic $V_i$ that estimates the agent's value function based on global information, and the prosocial component is a critic $Q_i^{SW}$ that estimates social welfare using COMA \cite{coma}. Again, neither selfish nor prosocial critics share the parameters. The decentralized policies $\pi_i^\oplus$ are trained on a combination of selfish and social advantages $A_i$ and $A_i^{SW}$ via PPO \cite{schulman2017proximal}. The combined advantage $A_i^\oplus$ is normalized over batch. To enhance exploration, we apply entropy regularization \cite{A3C}, annealed to 0 over the course of training. All weights of the networks use orthogonal initialization \cite{hu2020provable}. Finally, neither of the components utilizes experience replay.

In Eldorado, both agents receive global information as inputs. The state space is a vector with 28 elements. It includes statuses of food tiles, as well as characteristics of both agents, such as their coordinates, health points, resources, and actions taken in the previous turn. The action space consists of 10 possible options, which include 4 movement options, an option to pass, and an option to attack (combined with movement and passing). The only reward that each agent receives is $+1$ upon surviving for 1000 steps or $-1$ upon earlier termination.

In Harvest, each agent's local observation is restricted to a 15 by 15 part of the map, whereas the global state includes information about the whole 16 by 38 map. Both local and global states are 3-dimensional RGB images and are always preprocessed with a 6-channel CNN. The action space consists of 8 possible options, which include 4 movement options, 2 turn options, an option to pass, and an option to attack. The reward structure is the same as in the original implementation \cite{hughes2018inequity}: each agent receives $+1$ per collected apple, $-50$ for being attacked directly, and $-1$ for stepping into the fire left after an attack.

\begin{algorithm*}[]
\caption{BAROCCO for Q-learning framework}\label{alg:barocco_q}
\begin{algorithmic}

\STATE{\textbf{Initialize} Replay buffers $D$ and $D^{SW}$ for selfish and prosocial components}

\STATE{\qquad Networks $\theta_i$, $\theta_i^{SW}$, $\theta^{SW}$ that predict action-values $Q_i$, $Q_i^{SW}$, $Q^{SW}$}

\STATE{\qquad Hypernetwork $\theta^{H}$ that predicts weights of mixing network $\theta^{SW}$}

\STATE{\qquad Target networks $\overline{\theta}_i$}

\WHILE{True}

  \FOR{transition $t = 0\ldots T$}
  
    \STATE Sample weights in noisy layers, reduce exploration rate $\epsilon$
    
    \FOR{agent $i = 0\ldots N$}
    
      \STATE With probability $\epsilon$ sample random action $a_{i_t}$
      
      \STATE Otherwise, select $a_{i_t} = argmax_{a_{i_t}'} Q_i^\oplus\left(o_{i_t}, a_{i_t}'; \theta_i, \theta_i^{SW}\right)$
      
      \STATE \qquad where $Q_{i}^\oplus\left(o_{i_t}, a_{i_t}; \theta_i, \theta_i^{SW}\right) = (1 - \lambda) Q_{i}\left(o_{i_t}, a_{i_t}; \theta_i\right) + \lambda Q_{i}^{SW}\left(o_{i_t}, a_{i_t}; \theta_i^{SW}\right)$
    
    \ENDFOR

    \STATE Apply agents' actions, observe rewards and next state
    
    \STATE Store transitions $(\textbf{o}_t, \textbf{a}_t, \textbf{r}_t)$ to $D$, $(s, \textbf{o}_t, \textbf{a}_t, \textbf{r}_t)$ to $D^{SW}$
    
  \ENDFOR
  
  \STATE
  
  \FOR{agent $i = 0\ldots N$}
  
    \STATE Sample mini-batch of transitions $B_i$ from $D$ to update selfish action-value $Q_i$
  
    \FOR{transition $t = 0\ldots B_i$}
  
      \STATE $y_{i_t} = \overset{n}{\underset{k=0}{\sum}} \gamma^k r_{i_{t+k}} + \gamma^n Q_i\left(o_{i_{t+n}}, argmax_{a_{i_{t+n}}'} Q_i^\oplus(o_{i_{t+n}}, {a_{i_{t+n}}'}; \theta_i, \theta_i^{SW}); \overline{\theta}_i\right)$
      
    \ENDFOR
      
    \STATE Update $\theta_i$ via gradient descent on temporal difference loss $L_{TD}\left(Q_i(o_{i_B}, a_{i_B}; \theta_i), y_{i_B}\right)$
  
  \ENDFOR
  
  \STATE Periodically, copy weights of online networks $\theta_i$ to target networks $\overline{\theta}_i$
  
  \STATE
  
  \STATE Sample mini-batch $B$ from $D^{SW}$ to update prosocial action-values $Q_i^{SW}$ and $Q^{SW}$
  
  \FOR{transition $t = 0\ldots B$}
  
    \STATE Utilize global state via hypernetworks $\theta^{SW} = \theta^H(s_t)$
  
    \STATE $Q^{SW}\left(\textbf{o}_t, \textbf{a}_t; \theta_\textbf{i}^{SW}, \theta^{SW}, \theta^H\right) = \theta^{SW}\left(Q_{1}^{SW}(o_{1_t}, a_{1_t}; \theta_{1}^{SW}), \dots, Q_{N}^{SW}(o_{N_t}, a_{N_t}; \theta_{N}^{SW})\right)$
  
    \FOR{agent $i = 0\ldots N$}
  
      \STATE $y_{i_t} = r_{i_{t}} + \gamma Q_i\left(o_{i_{t+1}}, argmax_{a_{i_{t+1}}'} Q_i^\oplus(o_{i_{t+1}}, {a_{i_{t+1}}'}; \theta_i, \theta_i^{SW}); \overline{\theta}_i\right)$
      
    \ENDFOR
    
    \STATE $y_t^{SW} = SW\left(y_{1_t}, \dots, y_{N_t}\right)$
  
  \ENDFOR
  
  \STATE Update $\theta_\textbf{i}^{SW}$, $\theta^{SW}$, $\theta^{H}$ via gradient descent on $L_{TD}\left(Q^{SW}(\textbf{o}_B, \textbf{a}_B; \theta_\textbf{i}^{SW}, \theta^{SW}, \theta^H\right), y_{B}^{SW})$

\ENDWHILE


\end{algorithmic}
\end{algorithm*}

\begin{algorithm*}[h]
\caption{BAROCCO for Actor-Critic framework}\label{alg:barocco_ac}
\begin{algorithmic}

\STATE \textbf{Initialize} Critic networks $\theta_i$, $\theta_i^{SW}$ that predict values $V_i$, $Q_i^{SW}$

\STATE \qquad Actor networks $\psi_i$ that predict policies $\pi_i$

\WHILE{True}

  \FOR{transition $t = 0\ldots T$}
      
    \STATE Sample agents' actions $a_{i_t}$ from respective policies $\pi_i(o_{i_t}; \psi_i)$

    \STATE Apply agents' actions, observe rewards and next state
    
    \STATE Store transition $(s, \textbf{o}_t, \textbf{a}_t, \textbf{r}_t, s_{t+1}, \textbf{o}_{t+1})$ to batch $B$
    
  \ENDFOR
  
  \STATE Set $\pi_{old_i} = \pi_i$
  
  \FOR{mini-batch $b \in B$ }
  
  \FOR{agent $i = 0\ldots N$}
  
    \FOR{transition $t \in b$}
  
      \STATE $y_{i_t} = \overset{n}{\underset{k=0}{\sum}} \gamma^k r_{i_{t+k}} + \gamma^n V_i\left(s_{t+n} \mid a_{-i_{t+n}}; \theta_i\right)$
      
      \STATE $A_{i}\left(s_{i_t}, a_{i_t} \mid a_{-i_t}; \theta_i\right) = y_{i_t} - 
      V_i\left(s_{t} \mid a_{-i_t}; \theta_i\right)$
      
     \STATE $A_{i}^{SW}\left(s_{i_t}, a_{i_t} \mid a_{-i_t}; \theta_i^{SW}\right) =$  
      \STATE \qquad \qquad $Q_i^{SW}\left(s_{t}, a_{i_t} \mid a_{-i_t}; \theta_i^{SW}\right) - \sum_{a_i'} \pi_i(a_i' \mid o_{i_t}) Q_i^{SW}(s_t, a_i' \mid a_{-i_t}; \theta_i^{SW})$
      
      \STATE $\mathcal{R}_{i_t} = \frac{\pi_i(a_{i_t} | o_{i_t}; \psi_i)}
      {\pi_{old_i}(a_{i_t} | o_{i_t})}$
      
    \ENDFOR
      
    \STATE Update $\theta_i$ via gradient descent on temporal difference loss $L_{TD}\left(V_i(s_{b} \mid a_{-i_b}; \theta_i), y_{i_b}\right)$
    
    \STATE Update $\psi_i$ on PPO loss $L_{\pi_i}\left(A_{i}^\oplus\left(s_{i_b}, a_{i_b} \mid a_{-i_b}; \theta_i, \theta_i^{SW}\right)\right)$ with entropy regularization 
    \STATE \qquad where $A_{i}^\oplus\left(s_{i_b}, a_{i_b} \mid a_{-i_b}; \theta_i, \theta_i^{SW}\right) =$
    \STATE \qquad \qquad \qquad$(1 - \lambda) A_{i}\left(s_{i_b}, a_{i_b} \mid a_{-i_b}; \theta_i\right) + \lambda A_{i}^{SW}\left(s_{i_b}, a_{i_b} \mid a_{-i_b}; \theta_i^{SW}\right)$

  \ENDFOR
  \STATE $y_b^{SW} = SW\left(y_{1_b}, \dots, y_{N_b}\right)$
    \FOR{agent $i = 0\ldots N$}
    \STATE Update $\theta_i^{SW}$ via gradient descent on temporal difference loss:
    \STATE \qquad $L_{TD}\left(Q_i^{SW}(s_{i_b}, a_{i_b}\mid a_{-i_b}; \theta_i^{SW}), y_b^{SW}\right)$
      \ENDFOR
  \ENDFOR

\ENDWHILE

\end{algorithmic}
\end{algorithm*}

\bibliographystyle{icml2021}
\balance
\bibliography{Lawmaker}

\end{document}